\documentclass[letterpaper, 10 pt, journal]{ieeetran}

\IEEEoverridecommandlockouts                              

\usepackage{color}
\usepackage{soul}
\usepackage{CJKutf8}

\title{
   Autonomous Obstacle Removal for Excavators through Policy Learning with Particle Simulation
}

\author{Yuki Kadokawa$^{1}$, Sandro M. Alcantara Tacora$^{1}$, Taro Abe$^{2}$, \\Daisuke Endo$^{2}$, Genki Yamauchi$^{2}$, Takeshi Hashimoto$^{2}$, Takamitsu Matsubara$^{1}$
\thanks{
    This work was supported by JST Moonshot Research and Development, Grant Number JPMJMS2032. 
    $^{1}$ Nara Institute of Science and Technology, Nara 630-0192, Japan.
    $^{2}$ Public Works Research Institute, Ibaraki 300-2621, Japan.
}%
}

\usepackage{setspace}
\usepackage{comment}
\usepackage{subfigure}
\usepackage{booktabs}
\usepackage{iftex}
\ifPDFTeX
  \usepackage{graphicx}
\else
  \usepackage[dvipdfmx]{graphicx} 
\fi
\usepackage{amsmath}
\usepackage{amsfonts}
\usepackage{algorithmic}
\usepackage{siunitx}
\usepackage{bm}
\usepackage[space, compress]{cite}
\usepackage[ruled,vlined]{algorithm2e}
\usepackage{algorithm2e,setspace}
\usepackage{textcomp}
\usepackage{mathcomp}
\usepackage{tabularx}
\usepackage{here}
\usepackage[colorlinks=true, linkcolor=blue, citecolor=blue, urlcolor=blue]{hyperref}
\newcommand{\tabref}[1]{\hyperref[#1]{Table~\ref*{#1}}}
\newcommand{\equref}[1]{\hyperref[#1]{Eq.~(\ref*{#1})}}
\newcommand{\figref}[1]{\hyperref[#1]{Fig.~\ref*{#1}}}
\newcommand{\chapref}[1]{\hyperref[#1]{Section~\ref*{#1}}}
\newcommand{\algref}[1]{\hyperref[#1]{Algorithm~\ref*{#1}}}
\newcommand{\apperef}[1]{\hyperref[#1]{Appendix~\ref*{#1}}}

\begin{document}

\markboth{}%
{}

\maketitle

\begin{abstract}
Autonomous obstacle removal from the ground is an important earthwork task, but this is difficult to automate because an excavator must adapt its excavation trajectories over repeated cycles as soil-obstacle conditions change. Learning such state-dependent behavior requires a training environment that reproduces accumulated soil-obstacle interactions, including contact states, terrain deformation, and obstacle visibility. Accordingly, particle-based simulation is suitable for the relevant policy learning. However, particle simulation is computationally expensive, and repeated excavation cycles further increase the learning cost. We observe that the burial condition of an obstacle governs both task difficulty and simulation cost: deeper burial makes obstacle removal harder while also requiring more particles for accurate simulation. This observation motivates a burial-conditioned curriculum learning strategy. We propose a time-efficient sim-to-real policy learning framework in which the policy observes terrain and obstacle information from RGB-D measurements and then outputs a parameterized excavation trajectory; in this process, the simulator reproduces in a real-world excavator the same observation-action interface it uses under controllable burial conditions. The curriculum begins with shallow burial conditions and progressively increases burial depth while adjusting particle count, thus simultaneously controlling task difficulty and simulation cost. Experiments show that the proposed framework successfully learns an effective obstacle-removal policy, whereas baseline methods fail even after a full week of training. The proposed curriculum achieves effective performance within three days and achieves successful transfer to a real 12-ton excavator operating on open ground with various steel obstacles, thus demonstrating robust obstacle removal. A supplementary video is available at: \url{https://youtu.be/PycVGYh-cYs}
\end{abstract}


\section{Introduction}

In this paper, we aim to develop a policy learning framework for obstacle removal from the ground using an excavator. This task frequently arises in earthwork operations, such as excavation around buried pipelines and the removal of rocks and other obstacles \cite{survey_auto_excavation,survey_auto_excavation2}. In this task, excavation actions must be adaptively generated depending on the position and burial condition of a given obstacle, making it difficult to directly apply conventional frameworks developed for bulk soil excavation \cite{soil_rock_other_soil,excavator-sim2real}. In particular, due to the physical constraint that the excavator's bucket scoops soil as it advances, it is difficult to extract only the obstacle in a single excavation movement. Instead, the surrounding soil must be gradually removed before moving the obstacle. Consequently, the task requires repeated excavation cycles involving interactions with the soil, leading to a complex control problem characterized by strong state dependency and sequential decision-making \cite{rigid_clutter_excavation2021,tasknet2021,offline_rigid_excavation2023}.

\begin{figure}[t]
    \centering
    \includegraphics[width=0.99\columnwidth]{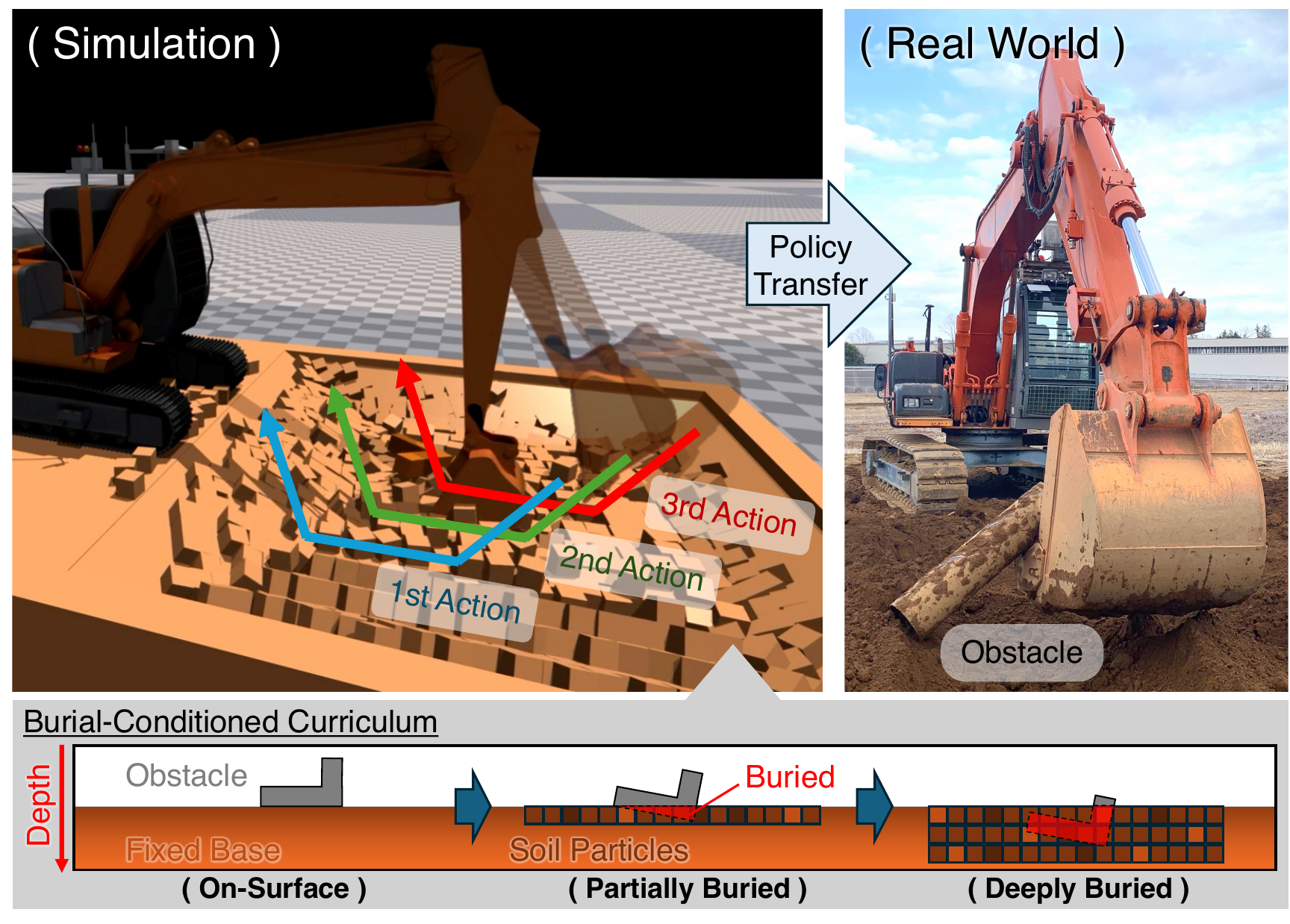}
    \caption{
        Overview of proposed sim-to-real policy learning framework for obstacle removal from ground using an excavator. This framework combines policy learning in particle simulation with real-world policy execution under a shared observation--action interface. During training, simulation settings are progressively organized according to burial condition of the obstacle so that task difficulty and simulation cost can be jointly controlled for time-efficient learning. Learned policy is then deployed on a real excavator through the same interface, providing direct sim-to-real transfer.
    }
    \label{fig:promotional_photo}
\end{figure}

To address such a control problem, previous works on soil excavation have adopted sim-to-real reinforcement learning approaches, where a policy is optimized through repeated interactions in simulation and then transferred to real-world systems \cite{excavator-sim2real,excavator-sim2real-geometric,vortex-DRL-sim2sim}. Since this training process requires many interaction samples, the soil simulation model is a central design choice for generating the samples efficiently and accurately. The simulation models used in earthwork sim-to-real learning can be broadly categorized into analytical models and particle-based models. Analytical models approximate soil using continuum, geometric, or force-based formulations, enabling efficient sample generation for bulk excavation and loading tasks \cite{soil_formula_1,soil_formula_2,auto_excavation_mpc}. However, obstacle removal requires resolving discontinuous local phenomena, including bucket--obstacle contact, soil--obstacle separation, local collapse, and progressive exposure of a rigid object, which are difficult to represent with approximations \cite{excavator-sim2real-geometric,rigid_clutter_excavation2021}. Particle simulation is thus more suitable for this task because it represents soil as discrete particles and explicitly computes interactions among the soil, the robot, and various obstacles \cite{particle_method_sim_meshFree,PRPD,isaacgym_excavation}.

However, the burial condition of a given obstacle affects not only task difficulty but also the computational cost of particle simulation. Burial conditions range from the obstacle being fully exposed to being partially buried or even barely observable, and each case requires the simulator to reproduce different soil--obstacle interactions and observation states. 
Deeper or less visible obstacles require more intricate soil-obstacle interactions to be reproduced and thus more surrounding soil particles for higher fidelity \cite{excavator_sim_particleLikeSim,rigid_clutter_excavation2021}.
Because particle simulation computes many particle--particle and particle--rigid-body interactions sequentially, increasing the particle count directly increases the computational cost per sample \cite{particle_method_sim_meshFree,PRPD,isaacgym_excavation}. As a result, collecting sufficient training samples across the full range of burial conditions becomes impractical in reinforcement learning.

To reduce computational costs, we focus on the observation that the burial condition of obstacles is not uniform but ranges from simply lying on the ground's surface to being deeply buried. In shallow burial conditions, interactions between the obstacle and the surrounding soil are limited, and only a small number of soil particles are required for accurate simulation. In contrast, deep burial conditions require both fuller soil interaction for simulation and more excavation steps to remove the surrounding soil before lifting the obstacle. Therefore, the burial condition conjoins two factors that are usually difficult to balance: task difficulty and required simulation fidelity. Shallow conditions are not only cheaper to simulate but also easier versions of the target removal task, whereas deeper conditions are both more difficult and more computationally expensive. This coupling makes the burial condition a physically meaningful curriculum variable, allowing training to progress from easy, low-cost stages to difficult, high-fidelity stages while jointly controlling learning difficulty and simulation cost \cite{CRL2,CRL}. Based on this insight, we hypothesize that a burial-conditioned curriculum can provide a more time-efficient approach to particle-simulation-based policy learning.

Motivated by this conjecture, we propose a time-efficient sim-to-real policy learning framework for obstacle removal from the ground using an excavator (\figref{fig:promotional_photo}). The framework first defines an observation--action interface that can be executed in the real world, where RGB-D measurements provide terrain and obstacle information and the policy outputs a parameterized excavation trajectory. We then construct a particle simulation environment that reproduces this interface while allowing the burial condition of the obstacle to be controlled. Then, we employ a burial-conditioned curriculum, which starts training from shallow, low-particle configurations and gradually increases burial depth and particle count. This design simultaneously controls task difficulty and simulation cost, improving sample and wall-clock training efficiency while ensuring direct deployment to a real excavator.
Experimental results in particle simulation with the burial-conditioned curriculum show that, while a conventional method achieves a roughly \SI{20}{\%} success rate after one full week of training, our method achieves over \SI{90}{\%} within three days. The learned policy is further validated in real-world experiments that demonstrate the stable operation of a 12-ton excavator on open ground with various steel obstacles.

The contributions of this paper can be summarized as follows.
\begin{itemize}
    \item We propose a framework for autonomous obstacle removal from the ground, formulating the problem as a state-dependent excavation task that requires repeated excavation cycles and sequential interaction with both soil and obstacles. The framework employs a trajectory-level action representation together with combined depth and obstacle-label observations.
    \item We propose a burial-conditioned curriculum that exploits the burial condition of the obstacle as a common factor governing both task difficulty and simulation cost, and we show that progressively increasing particle count reduces wall-clock training time while maintaining high task performance.
    \item We verify the practical feasibility of the proposed framework through real-world excavator experiments, demonstrating successful sim-to-real transfer and clarifying the remaining efficiency gap from human experts.
\end{itemize}

\section{Related Works}
    \label{s:related_works}

    \subsection{Learning Frameworks for Excavation}
        This section reviews how excavation behavior has been learned or adapted using real-world trials, analytical or continuum-model simulations, and particle simulations. Moreover, among such studies, it positions our framework for obstacle removal from the ground.

        \textbf{Real-world learning:} Previous works have learned or adapted excavation skills directly in real environments \cite{rock_excavation_humanBetter,rock_excavation_complex_soil,auto_excavation_mpc}. These methods either update parameterized bucket trajectories from repeated digging trials or estimate bucket--soil interaction models from measured excavation data to improve low-level controllers \cite{rock_excavation_realOnly_GPmodelNoPolicyLearning,soil_excavation_realOnly_GPmodelMPCPolicyLearning}. Although real interaction data can refine specific excavation motions or force-related controllers, real-world trial-and-error efforts using excavators is costly, limiting these approaches to narrow settings such as a fixed digging motion or a learned interaction model. Therefore, applying such frameworks to obstacle removal is difficult, since the excavation action must be reselected after each trial as the relative states of obstacle and surrounding soil change; furthermore, collecting sufficiently diverse samples for such state-dependent decisions is impractical with a real excavator.

        \textbf{Simulation-based learning with analytical and continuum models:} To avoid real-world sampling costs, many studies learn excavation behavior in simulation for bulk soil excavation, loading, and trajectory generation \cite{ImitationExcavation,DeepImitation_LSTM_excavation_soil,soil_rock_other_soil}. These works commonly use analytical, geometric, or continuum-based models to predict a bucket's path or evaluate candidate paths and force-related parameters for each digging cycle \cite{soil_formula_2,excavator-sim2real,FEM_excavation}. Because such models approximate excavation forces and terrain deformation with simplified equations or continuum formulations, they are computationally efficient and suitable for large-scale training when the target is mainly soil and generally uniform digging behavior can be reused while updating the terrain state \cite{excavator-sim2real-geometric,rigid_clutter_excavation2021,fragmented_rocks_mbrl2022}. However, this formulation is difficult to apply to obstacle removal, where a persistent rigid obstacle causes intermittent bucket--obstacle contact, soil--obstacle separation, local collapse, and changes in visibility after each excavation attempt. Since these discontinuous local interactions are not explicitly tracked, analytical or continuum models alone are insufficient for learning state-dependent excavation actions over repeated trials of removal.

        \textbf{Simulation-based learning with particle simulation:} Particle simulations represent soil as discrete particles or material points and explicitly compute their interactions with rigid bodies and robots \cite{jiang2016mpm,particle_method_sim_meshFree,isaacgym_excavation}. By resolving local motion, contact, and friction, these methods reproduce bucket--soil collision, soil--obstacle contact and separation, local collapse, and progressive exposure of a buried object within a unified simulation process \cite{PRPD,FEM_excavation}. This level of expressiveness makes particle simulation suitable for obstacle removal from the ground, but it also introduces a substantial computational burden because many particle--particle and particle--rigid body contacts must be evaluated repeatedly. Moreover, deeper burial and more complex local deformation require more particles for representing more complex soil-obstacle interactions, further increasing the training cost. Accordingly, obstacle removal requires both an expressive simulator and a policy learning framework that remains practical within its permitted computational cost.

        \textbf{Positioning of this study:} This paper addresses obstacle removal from the ground as a sequential policy-learning problem, where each excavation changes the obstacle position, burial condition, terrain, visibility, and contact state, requiring the next trajectory to be selected accordingly. Unlike real-world learning approaches, we do not rely on collecting diverse trial-and-error samples with an actual excavator; unlike analytical or continuum-model simulations, we use particle simulation to represent local soil--obstacle interactions. The main challenge is thus not only to obtain an expressive simulator but also to make policy learning feasible despite the computational cost. We tackle this challenge by exploiting the staged structure of burial conditions to implement training in which task difficulty and particle-simulation cost are jointly controlled.

    \subsection{Training Efficiency in Particle Simulation}
        This section reviews approaches to reducing the wall-clock cost of policy learning with particle simulation, focusing on sample reduction, simulator acceleration, and staged training design.

        \textbf{Reducing required samples:} Particle simulation makes each rollout expensive, so imitation learning and related low-sample strategies have been used to reduce the number of interactions needed for training \cite{ImitationExcavation,DeepImitation_LSTM_excavation_soil,IL}. These methods learn from demonstrations or supervisory signals instead of relying solely on exhaustive exploration \cite{imitation_learning,excavator_complex_vortex}. Reinforcement learning with particle-based simulators has also been combined with sample-efficient techniques such as particle-resolution transfer and policy reuse across different simulation settings \cite{vortex-DRL-sim2sim,isaacgym_excavation,offline_rigid_excavation2023,PRPD,cpd}. These approaches can reduce exploration burden and improve data reuse, but they do not directly reduce the computational cost of each simulated interaction.

        \textbf{Reducing simulator cost:} Another direction improves efficiency at the simulator level. Representative methods of this type apply coarser or aggregated particle representations, transfer policies across particle resolutions, restrict expensive computation to strongly interacting regions, or exploit GPU-based execution and optimized granular simulators \cite{PowderWeighingSim2Real,PRPD,cpd}. These methods share the goal of reducing the computational burden of particle simulation, especially by lowering the effective particle count or execution time. Consequently, they reduce per-step calculation time and increase the number of samples collected within a fixed wall-clock budget. However, simulator-level acceleration alone cannot specify when lower or higher simulation fidelity should be used to cope with changing task conditions during learning.

        \textbf{Staged training design:} Curriculum learning improves training efficiency by starting from easier settings and gradually transferring the policy to more difficult ones \cite{CRL,CRL2,CRL_reward1}. Related studies also use curricula over simulation fidelity, for example by increasing particle resolution during learning \cite{PRPD,cpd}. The results of these works show that time efficiency can be improved not only by faster simulation but also by organizing the training sequence. Nevertheless, task difficulty and simulation fidelity are typically scheduled as separate factors, rather than being tied to a physical variable that determines both.

        \textbf{Positioning of this study:} In the above taxonomy, our framework combines staged training design with simulator-cost reduction. Similar to simulator-level acceleration methods, we reduce particle-simulation costs by avoiding unnecessarily high particle counts in stages where lower fidelity is sufficient. Similar to curriculum learning, we organize training from easier to harder settings. The difference is that both choices are governed by the same physical variable: The burial condition orders the obstacle-removal task by difficulty while also indicating the particle count required at each stage. Therefore, in the proposed framework, task difficulty and simulation cost are jointly scheduled, rather than treating curriculum design and simulator-cost reduction as separate techniques.

\begin{figure}[t]
    \centering
    \includegraphics[width=0.99\columnwidth]{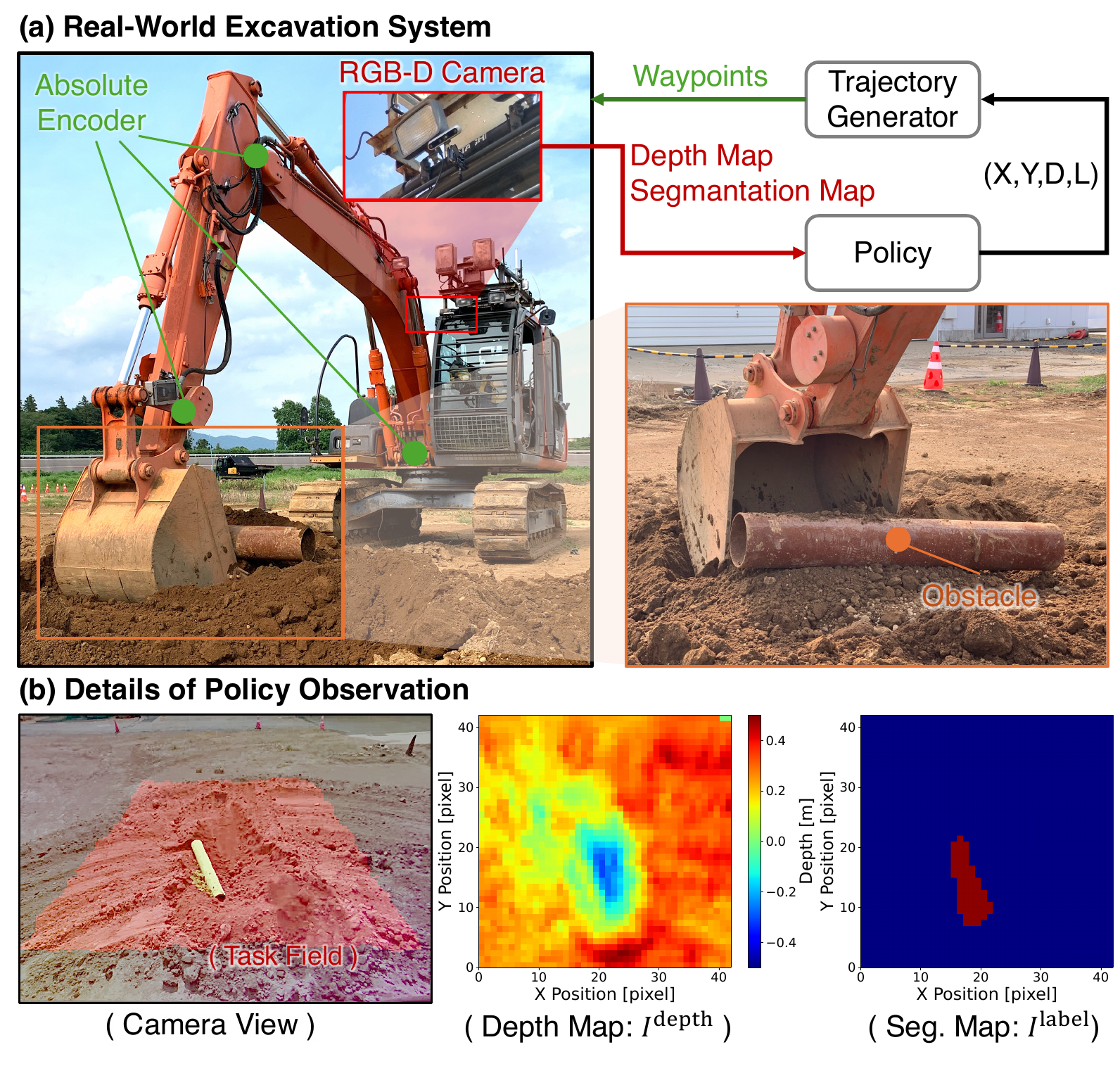}
    \caption{
        Overview of real-world observation--action interface. \textbf{(a)} Real-world excavation system, where an RGB-D camera observes the task field and joint absolute encoders are used for low-level control. Policy output $[X,Y,D,L]$ is converted into a bucket trajectory executed by the onboard controller. \textbf{(b)} Details of policy observation, where RGB-D measurements are processed into a depth map $I^{\mathrm{depth}}$ and an obstacle-label map $I^{\mathrm{label}}$.
    }
    \label{fig:real_setup}
\end{figure}

\section{Proposed Framework}

    This section presents the proposed sim-to-real policy learning framework for autonomous obstacle removal with an excavator. The framework is designed to satisfy three requirements: (1) obstacle removal should be formulated as a sequential decision-making problem requiring repeated excavation cycles, (2) the learned policy should be executable on a real-world excavator through a shared control interface, and (3) particle-based simulation training should maintain computational efficiency. In the following, we describe the policy interface, the real-world excavator system, and the particle-simulation training process in turn.

        \textbf{Task Assumption:}
            We consider a practical excavation scenario in which an excavator, in the course of ongoing digging, removes obstacles embedded in the ground. The proposed controller is activated when an obstacle becomes partially visible, for example when it appears on the ground's surface or is detected within the working area. Because practical excavation is typically continued with the bucket rather than specialized equipment, the obstacle must be removed through interaction with the surrounding soil rather than direct grasping. Owing to the physical constraints of excavation, it is usually impossible to complete this removal in a single bucket motion, thus requiring repeated excavation cycles to gradually displace the obstacle.

        \textbf{Policy Execution Process:}
            To perform autonomous obstacle removal, the policy operates in an iterative perception-action loop. At each step, the excavator observes the current field state, generates a parameterized excavation trajectory, and executes the motion. After execution, the bucket is lifted and its contents are discharged either to the obstacle-placement point when an obstacle is detected or to the soil-placement point otherwise. This cycle is repeated until the obstacle is completely removed from the task field.

    \subsection{Learnable Policy Interface}
    \label{sec:proposed_task_definition}

        This subsection defines the observation--action interface shared by the particle simulation and the real-world excavator. Instead of learning low-level joint commands, the policy predicts a compact trajectory-level action from visual observations. This design preserves the task-relevant information needed for sequential obstacle removal while keeping the exploration space small enough for efficient learning and direct sim-to-real deployment.
        
        \subsubsection{Action Design}
        \label{action_design}
            Each action is represented by a four-dimensional vector:
            \begin{equation}
                a = [X, Y, D, L]^\top,
            \end{equation}
            where $(X, Y)$ specifies the excavation position from the excavator's base frame, $D$ is the excavation depth, and $L$ is the drag distance. These parameters are converted by a deterministic trajectory generator into an executable bucket motion. Internally, the generated trajectory is organized by the approach, penetration, dragging, lift-out, and removal phases. Detailed definitions of this trajectory generator are provided in \apperef{sec:trajectory_parameterization_details}. The policy predicts only task-relevant excavation parameters, while the remaining motion structure is fixed by geometric rules; therefore, the action space remains low-dimensional and can be shared across simulation and real execution.

        \subsubsection{Observation Design}
            The observation is defined as
            \begin{equation}
                s_t = \left(I^{\mathrm{depth}}_t, I^{\mathrm{label}}_t\right),
            \end{equation}
            where $I^{\mathrm{depth}}_t$ is a depth map of the terrain and $I^{\mathrm{label}}_t$ is a segmentation map of the obstacle region. The depth map provides terrain geometry, whereas the label map provides obstacle location and extent. Using the same two-modal representation in both environments allows the learned policy to operate under a common interface while retaining the information needed for state-dependent excavation.

    \subsection{Real-World Excavator System}
    \label{sec:proposed_real_execution}

        This subsection describes the real excavator system that functions as the deployment target of the shared policy interface.

        \subsubsection{System Overview}
            As shown in \figref{fig:real_setup}, the excavator is equipped with an RGB-D camera for observing the task field and joint-mounted absolute encoders for controlling the swing, boom, arm, and bucket joints. Observations are acquired when the bucket is lifted to make the task field visible before each excavation cycle. Detailed perception and low-level execution pipelines are provided in \apperef{sec:real_execution_details}.

        \subsubsection{Observation and Action Execution}
            The real system constructs the observation pair $\left(I^{\mathrm{depth}}, I^{\mathrm{label}}\right)$ from RGB-D measurements and uses the policy output $a=[X,Y,D,L]^\top$ as a parameterized excavation command. The observation is converted into the fixed-resolution input used by the policy, and the command is transformed into an executable bucket trajectory tracked by the onboard controllers. Detailed definitions are provided in \apperef{sec:real_execution_details}.

\begin{figure}[t]
    \centering
    \includegraphics[width=0.99\columnwidth]{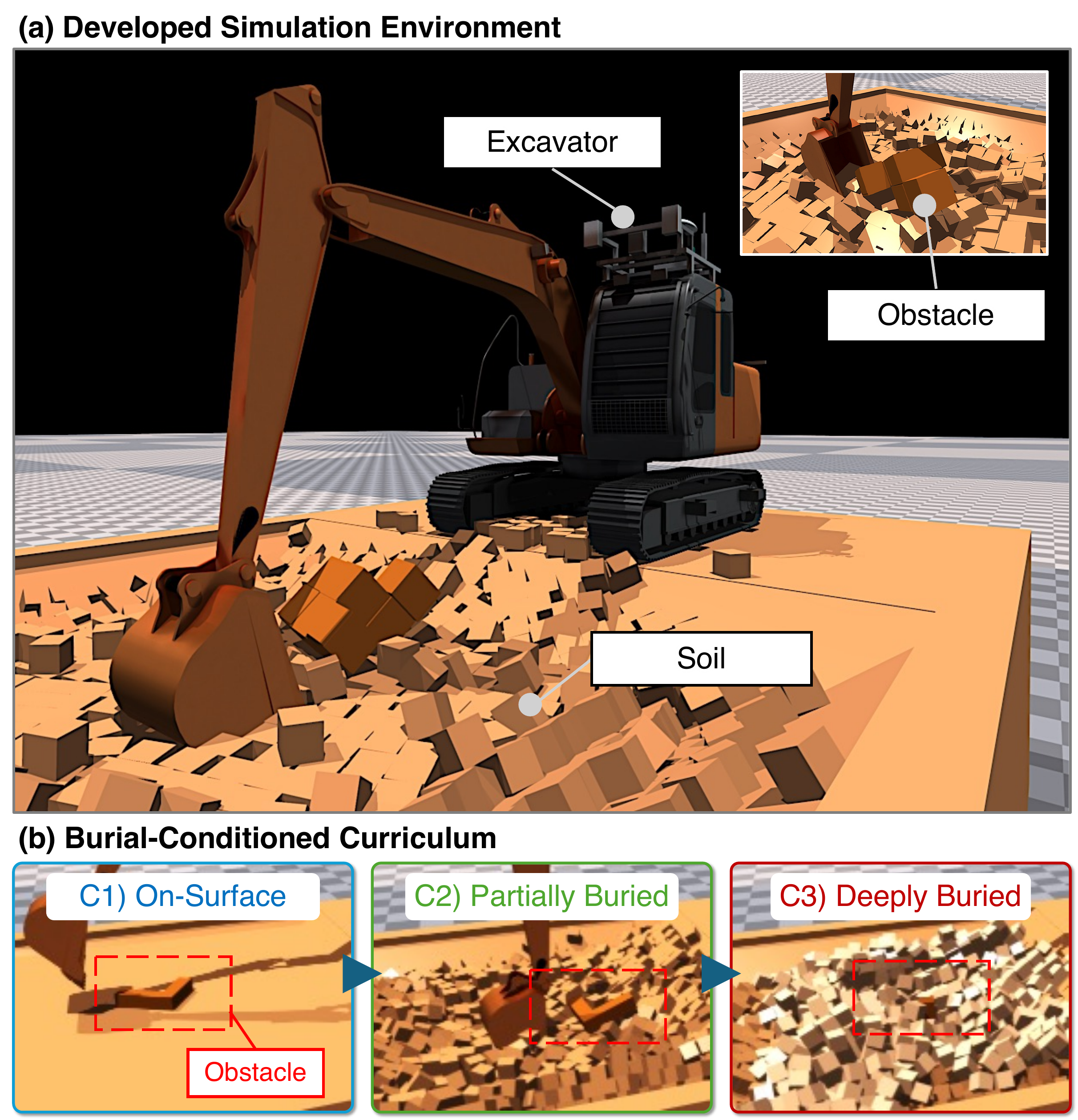}
    \caption{
        Overview of policy learning framework in particle simulation. \textbf{(a)} Developed simulation environment for obstacle removal, which models interactions among an excavator, particle-based soil, and various obstacles on or in the ground. \textbf{(b)} Buried-conditioned curriculum, where the obstacle setting progresses from On-Surface to Partially Buried and Deeply Buried (C1--C3) as the number of soil particles increases accordingly.
    }
    \label{fig:fig2}
\end{figure}

    \subsection{Time-Efficient Learning in Particle Simulation}
    \label{sec:proposed_simulation_learning}

        This subsection describes the particle-simulation training environment built to reproduce the real excavator system under the shared interface. We then introduce the burial-conditioned curriculum that adjusts particle count according to both task difficulty and simulation cost, followed by the reinforcement-learning and domain-randomization setup used to learn a transferable policy.

        \subsubsection{Simulation Environment}
    
            As illustrated in \figref{fig:fig2}, the simulation environment mirrors the real system by maintaining the same observation--action interface and controlled excavator joints. The excavator is represented by the swing, boom, arm, and bucket joints, and the virtual camera generates the depth and obstacle-label maps used by the policy. The terrain is modeled as a hybrid representation consisting of a global height field and particles confined to the task field, allowing local soil--obstacle interaction to be reproduced while limiting computational cost. Obstacles are modeled as rigid bodies placed on or in the ground according to the target scenario. Detailed descriptions of the virtual camera and low-level control interface are provided in \apperef{sec:simulation_dr_details}.

        \subsubsection{Burial-Conditioned Curriculum}
            To improve training-time efficiency, we organize learning according to the burial condition of the obstacle ( \figref{fig:fig2}). In this task, burial depth affects both task difficulty and computational cost: More deeply buried obstacles require greater soil--obstacle interaction and thus more particles. The proposed curriculum exploits this relation by starting from easier and cheaper settings and then progressively increasing particle count and burial complexity.

            The curriculum consists of three stages, denoted as C1, C2, and C3. In C1, the environment contains no soil particles, with the obstacle simply placed on a flat surface; ground contact by the bucket is disabled to enable obstacle interaction. In C2, a minimal set of soil particles is introduced to form a flat terrain with the obstacle on top. In C3, multiple soil clusters are generated to create diverse and more complex burial conditions. Training advances to the next stage after the current policy reaches a predefined target success rate. This progressive schedule allows the policy to first acquire coarse excavation behavior in computationally inexpensive settings and then refine it under richer interaction dynamics, thus reducing total training time while maintaining performance in high-fidelity environments.

        \subsubsection{Policy Learning and Domain Randomization}
    
            After specifying the simulation environment and curriculum schedule, the policy is trained in a standard reinforcement-learning framework with domain randomization under the shared observation--action interface defined in \chapref{sec:proposed_task_definition}. Formally, the obstacle-removal task is modeled as a Markov decision process (MDP) $\langle \mathcal{S}, \mathcal{A}, p, r, \gamma \rangle$, where $\mathcal{S}$ denotes the observation space, $\mathcal{A}$ the action space, $p$ the transition dynamics, $r$ the reward function, and $\gamma$ the discount factor. The policy $\pi(a_t\,|\,s_t)$ is optimized to maximize the expected discounted return
            \begin{equation}
                J(\pi) = \mathbb{E}_{\pi}\!\left[\sum_{t=0}^{T} \gamma^{t} r(s_t, a_t)\right].
            \end{equation}
            In our problem, the observation is $s_t = (I^{\mathrm{depth}}_t, I^{\mathrm{label}}_t)$, the action is $a_t = [X_t, Y_t, D_t, L_t]^\top$, and the transition $p$ is induced by the excavator--soil--obstacle interactions produced after executing the generated excavation trajectory in simulation. 
            
            The reward is defined as
            \begin{equation}
                r(s_t, a_t) = \omega_{\mathrm{success}} r^{\mathrm{success}}_t + \omega_{\mathrm{distance}} r^{\mathrm{distance}}_t.
            \end{equation}
            Here, $\omega_{\mathrm{success}}$ and $\omega_{\mathrm{distance}}$ are scaling coefficients for the success and distance terms, respectively. The success term $r^{\mathrm{success}}_t$ is $0$ when the obstacle has been removed from the task field and $-1$ otherwise, providing a sparse penalty until task completion. The distance term $r^{\mathrm{distance}}_t$ provides dense reward shaping by assigning a larger penalty when the bucket moves farther from the obstacle, thereby promoting actions that approach and manipulate the obstacle even before removal is achieved. This term is computed as the negative average distance between the bucket reference point and the obstacle center, $r^{\mathrm{distance}}_t = -|\mathcal{K}_t|^{-1}\sum_{k \in \mathcal{K}_t}\|\mathbf{b}_{t,k}-\mathbf{o}_{t,k}\|_2$, where $\mathbf{b}_{t,k}$ and $\mathbf{o}_{t,k}$ denote the bucket reference point and obstacle center at sampled waypoint $k$, respectively. In the implementation, the waypoint set $\mathcal{K}_t$ corresponds to the segment from $\mathbf{p}_3$ to $\mathbf{p}_4$, whose trajectory definitions are provided in \apperef{sec:trajectory_parameterization_details}. To improve robustness against the reality gap, domain randomization \cite{DR-DRL-LSTM,cpd} is applied at the beginning of each training episode over three groups of factors: soil and terrain properties, camera viewpoint and depth noise, and obstacle properties and shapes. Detailed randomization ranges and the obstacle-shape generation procedure are provided in \apperef{sec:simulation_dr_details}.

\section{Experiments}
\label{s:exp}

    This section presents simulation and real-world experiments that evaluate the proposed framework. The primary evaluations focus on policy learnability, wall-clock training efficiency, and sim-to-real transferability; additional analyses and ablation studies are provided in \apperef{sec:simulation_fidelity_analysis}, \apperef{sec:trajectory_design_analysis}, \apperef{sec:policy_learning_necessity}, and \apperef{sec:domain_randomization_contribution}.
    
    The main research questions addressed by these experiments are as follows:
    \begin{list}{}{%
        \setlength{\labelwidth}{3.3em}%
        \setlength{\labelsep}{0.35em}%
        \setlength{\leftmargin}{3.65em}%
        \setlength{\itemindent}{0pt}%
        \setlength{\listparindent}{0pt}%
        \setlength{\parsep}{0pt}%
        \setlength{\itemsep}{0.2em}%
        \setlength{\topsep}{0.2em}%
        \setlength{\partopsep}{0pt}%
    }
        \item[\textbf{(RQ1)}] Can the proposed framework learn an effective obstacle-removal policy for a state-dependent excavation task requiring repeated excavation cycles in comparison with existing training strategies (\chapref{sec:policy_learnability_comparison}) and key action/observation ablations (\chapref{sec:action_observation_analysis})?
        \item[\textbf{(RQ2)}] Does the proposed burial-conditioned curriculum improve wall-clock training efficiency by progressively increasing particle count during learning? (\chapref{sec:calculation_time_comparison})
        \item[\textbf{(RQ3)}] Can the learned policy transfer from the training simulator to both another simulator used for evaluation and a real excavator? (\chapref{sec:policy_transferability})
        \item[\textbf{(RQ4)}] How does the transferred policy compare with human experts in task-completion time, and which execution phases explain the difference? (\chapref{sec:task_completion_time_comparison})
    \end{list}

\subsection{Common Settings}
    This section describes only the experimental conditions shared across the main evaluations. Unless otherwise noted, all methods use the same excavator platform, perception device, observation--action interface, low-level controller, reward design, and reinforcement-learning algorithm; they primarily differ only in the component being evaluated. The numerical values of these shared settings are collected in \apperef{sec:experimental_parameters}, while experiment-specific settings are described in the corresponding subsections. A hydraulic excavator (Hitachi ZX120) is used as the robotic platform, and an RGB-D camera (Intel RealSense D435) is used for perception. The policy outputs the four excavation parameters introduced in \chapref{action_design}, which are converted into a predefined bucket trajectory by the common trajectory generator used in both simulation and real execution. Detailed definitions of the control points and generator parameters are given in \apperef{sec:trajectory_parameterization_details}. 
    For policy learning, we use Randomized Ensembled Double Q-Learning (REDQ) \cite{REDQ} in all comparisons. 
    A trial is regarded as successful if the obstacle is removed from the task field within the predefined action horizon.

\subsection{Comparison of Methods}
    \label{sec:comparing_methods}

    We compared the proposed framework with its ablation variants and related curriculum-based training strategies to verify the effectiveness of each design component. Unless otherwise noted, all methods share the same observation space, reward function, learning algorithm, and low-level controller, and they differ only in the training schedule or action representation.
    
    The compared methods are summarized as follows.
    \begin{itemize}
        \item Only C1: The policy is trained exclusively in Curriculum 1, where no soil particles are included and the task environment is simplified.
        \item Only C2: The policy is trained only in Curriculum 2, where a flat terrain with soil particles is considered, without progression from simpler environments.
        \item Only C3: The policy is trained directly in Curriculum 3, which includes complex terrain with soil particles and partially buried obstacles, without any curriculum progression. This configuration is equivalent to conventional domain-randomized RL performed only in the target environment.
        \item Only C3 (CDR) \cite{CRL}: The policy is trained only in Curriculum 3 with curriculum-based domain randomization (CDR) applied, but without progressive transitions from C1 and C2. In this variant, the domain-randomization range is expanded monotonically as a function of the number of collected samples while the task setting itself remains fixed to C3.
        \item Only C3 (Trj.): The policy is trained only in Curriculum 3 using a waypoint-based direct trajectory representation instead of the proposed four-parameter action representation. This baseline directly predicts intermediate excavation waypoints; its detailed implementation is discussed in \apperef{sec:trajectory_baseline_details}. Unless otherwise noted, this variant uses five intermediate points (``Trj. (5)'') as the default waypoint-based baseline; the effect of increasing the number of intermediate points is analyzed separately in \apperef{sec:trajectory_design_analysis}.
    \end{itemize}

\subsection{Policy Learnability Compared with Existing Methods}
\label{sec:policy_learnability_comparison}

    \subsubsection{Settings}
        This section evaluates whether the proposed framework improves policy learnability compared with other training approaches. We compare ``Ours'' with the representative baselines described in \chapref{sec:comparing_methods}.

    \subsubsection{Results}
\begin{figure}[t]
    \centering
    \includegraphics[width=0.99\columnwidth]{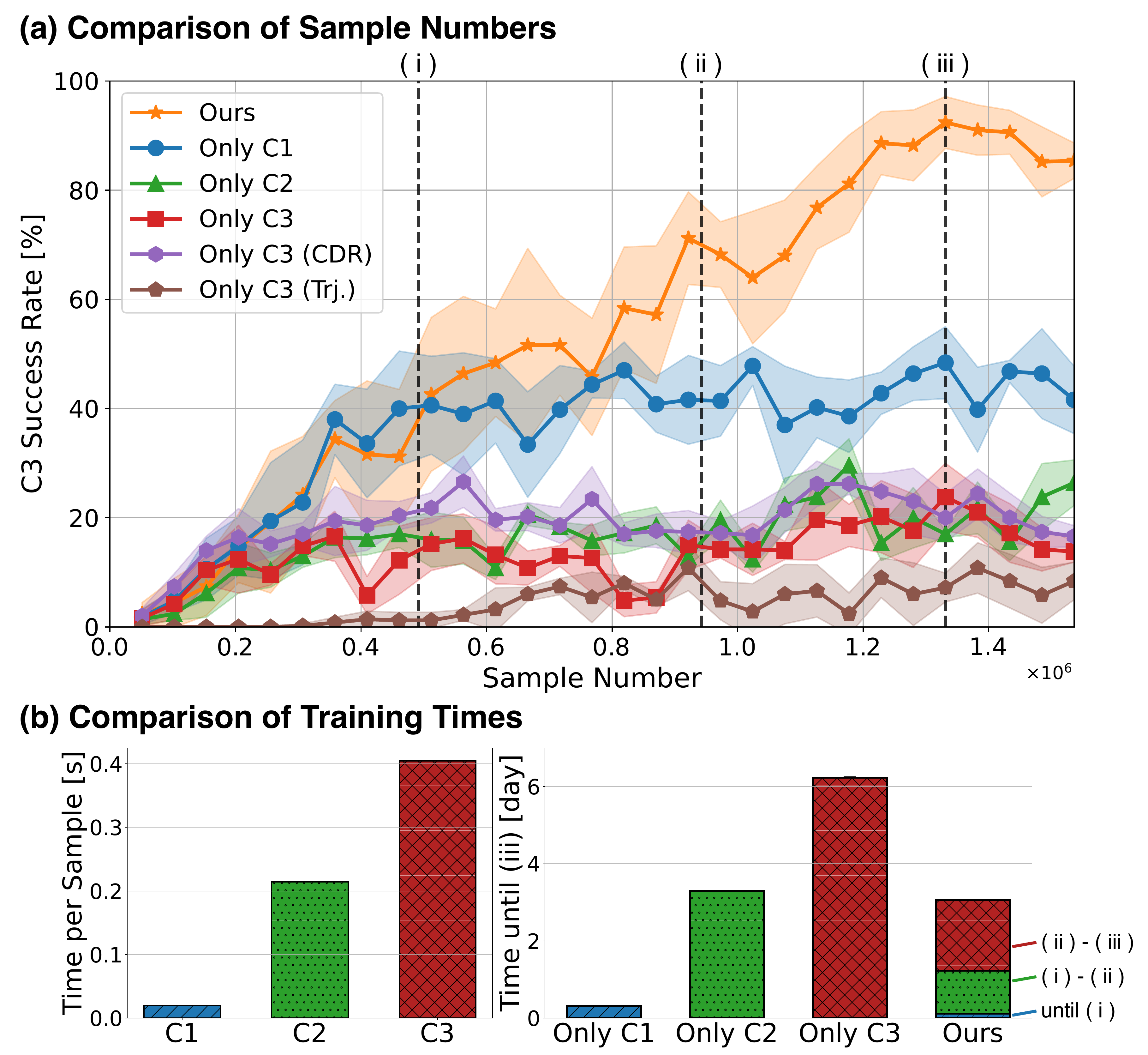}
    \caption{
        Comparison of sample efficiency and training time.
        \textbf{(a)} Learning curves. ``Only C1''--``Only C3'' denote training only in the corresponding curriculum. Markers (i)--(iii) indicate when ``Ours'' finishes C1, finishes C2, and reaches peak performance in C3. ``Only C3 (CDR)'' expands the domain-randomization range monotonically with the number of collected samples, and ``Only C3 (Trj.)'' uses the ``Trj. (5)'' setting. For C3, success rate is evaluated over 100 rollouts at each point. Each curve shows mean and standard deviation over five experiments.
        \textbf{(b)} Training time. Bars for C1--C3 show time required to collect a single action sample. ``Time until (iii)'' compares total wall-clock time required for ``Ours'' to reach (iii) with time required by each curriculum for the same number of samples.
    }
    \label{fig:learning_design}
\end{figure}

        As shown in \figref{fig:learning_design} (a), the proposed framework consistently achieves higher success rates, exceeding \SI{90}{\%}, whereas the compared baselines remain at about \SI{30}{\%} or below. These results indicate that neither direct training only in the final curriculum nor simple expansion of the domain-randomization range is sufficient for learning this task.
        Among the baselines, ``Only C3 (Trj.)'' learns especially slowly, suggesting that direct waypoint prediction makes exploration more difficult than the proposed trajectory parameterization. Although ``Only C3 (CDR)'' is more sample-efficient than ``Only C3,'' its performance still saturates far below that of ``Ours.'' These results suggest the effectiveness of combining the burial-conditioned curriculum with the proposed parameterized action representation.

\subsection{Validation of Action and Observation Designs}
\label{sec:action_observation_analysis}

    \subsubsection{Settings}
        To augment the above comparison with existing methods, this section examines how the proposed policy performs well by analyzing the contribution of the action and observation designs to task performance. We compare the full proposed framework (``Ours'') with ablation variants that modify either the action representation or the observation design. The concrete ablation settings are shown in \figref{fig:compare_task_design}.

    \subsubsection{Results}
\begin{figure}[t]
    \centering
    \includegraphics[width=0.99\columnwidth]{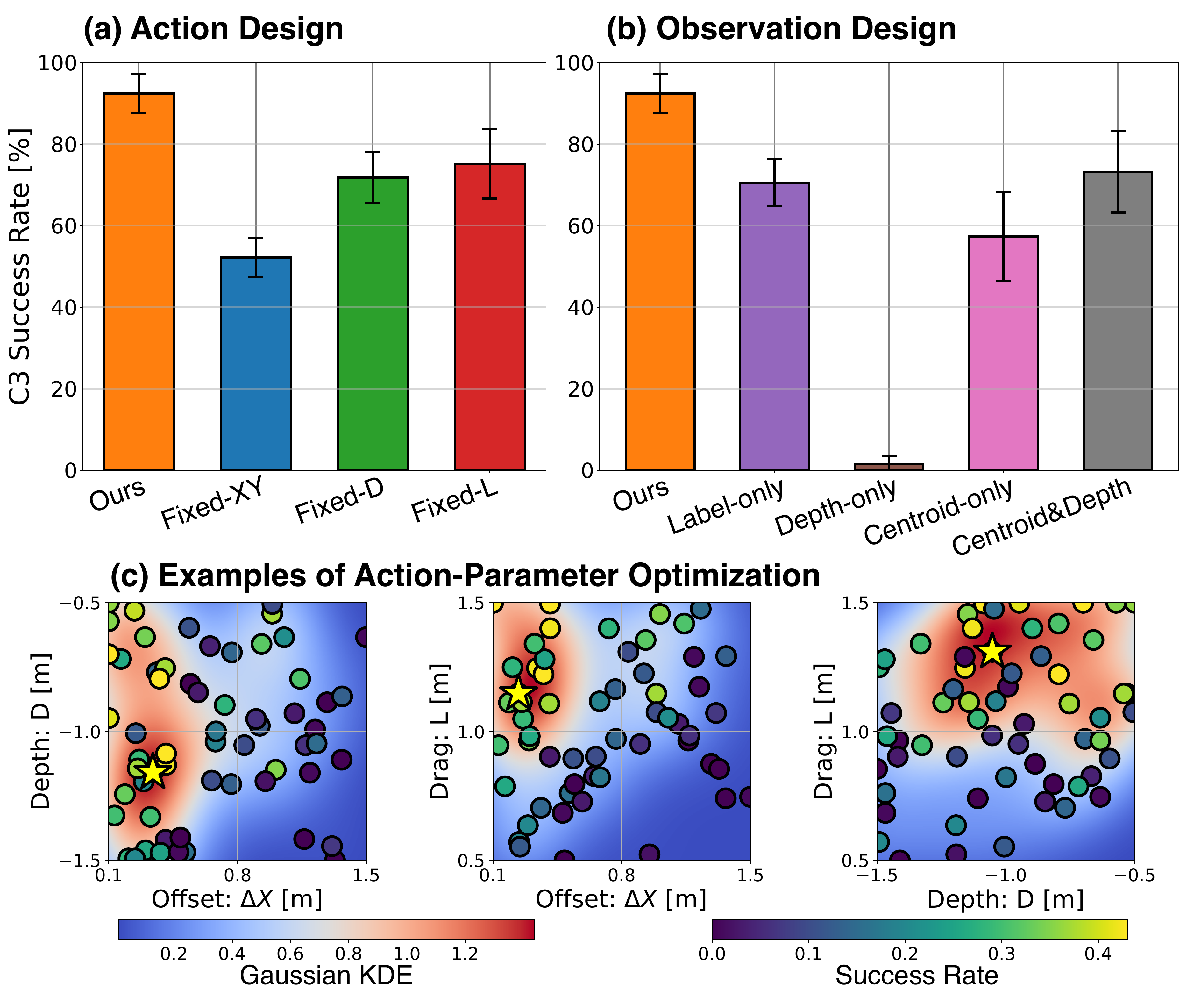}
    \caption{
        Comparison of action and observation designs.
        \textbf{(a)} Action-design ablation. Proposed framework predicts four action parameters $[X,Y,D,L]^\top$. In each ablation, a part of the action is replaced with fixed parameters optimized by Bayesian optimization. For fixed-XY setting, excavation position is defined as $(x+\Delta X,y)$ using obstacle centroid $(x,y)$ and an optimized insertion offset $\Delta X$.
        \textbf{(b)} Observation-design ablation. Proposed framework uses both terrain depth map and obstacle-label map. Ablations compare variants using only obstacle-label map, only terrain depth map, only obstacle centroid, and obstacle centroid combined with terrain depth map.
        \textbf{(c)} Example of Bayesian optimization used to determine fixed action parameters for centroid-based naive controller. Star marker denotes parameter setting achieving the highest success rate in this trial.
        For designs that do not exceed a \SI{90}{\%} success rate within a curriculum stage, training is capped at 2000 episodes per curriculum before transitioning to the next stage. Each bar shows mean and standard deviation over five experiments.
    }
    \label{fig:compare_task_design}
\end{figure}

        \figref{fig:compare_task_design} summarizes the ablation results for the proposed action and observation designs. For the action design, replacing policy-generated trajectory parameters with fixed parameters optimized by Bayesian optimization consistently degrades performance. The largest drop occurs when the XY parameters are fixed from the obstacle centroid, reducing the success rate by approximately \SI{40}{\%}, while fixing $L$ or $D$ also lowers performance by more than \SI{10}{\%}. These results indicate that state-dependent generation of trajectory-level actions is important for completing this task. As for the observation design, the ``Depth-only'' variant performs poorly, with success rates below \SI{10}{\%}, showing that the obstacle state cannot be inferred reliably from terrain geometry alone. The ``Label-only'' variant performs better but is still approximately \SI{20}{\%} below ``Ours,'' and replacing the full label map with obstacle centroid coordinates causes a further drop of about \SI{20}{\%}. Taken together, these results show that the policy benefits from combining terrain depth with spatially dense obstacle-label observations.

\begin{figure*}[t]
    \centering
    \includegraphics[width=1.99\columnwidth]{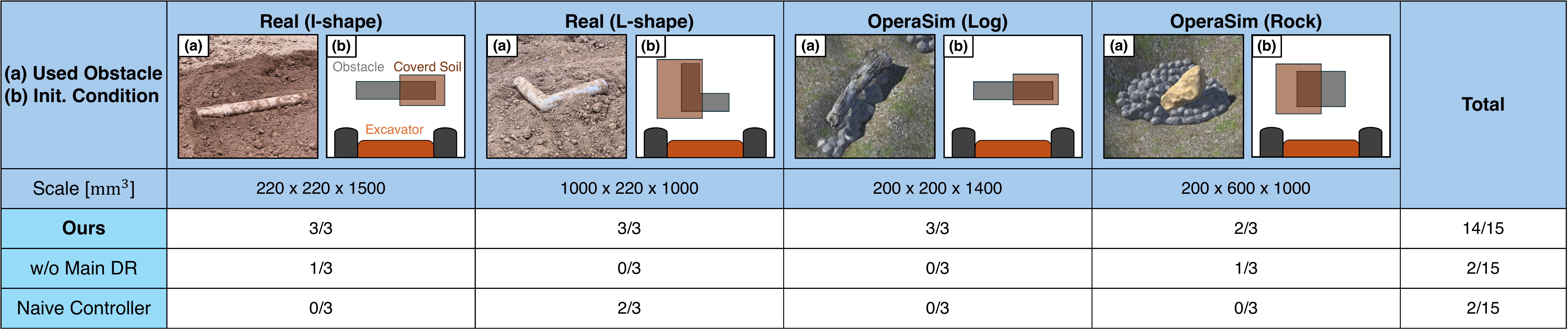}
    \caption{
        Comparison of task success rates in sim-to-sim and sim-to-real experiments. In real-world environment, I-shaped and L-shaped obstacles are used, whereas logs and rocks are used in the evaluation simulator. Snapshots of these obstacles are shown in first row. Each method is evaluated using three independently trained policies, and each reported value indicates number of successful trials. Initial obstacle position is set at $x = 5$ and $y = 0$. Detailed obstacle orientations and soil-cover conditions are shown in second row. Terrain is not leveled in advance and thus contains randomly distributed mounds and holes.
    }
    \label{fig:task_achievement_real}
\end{figure*}

\begin{figure*}[t]
    \centering
    \includegraphics[width=1.99\columnwidth]{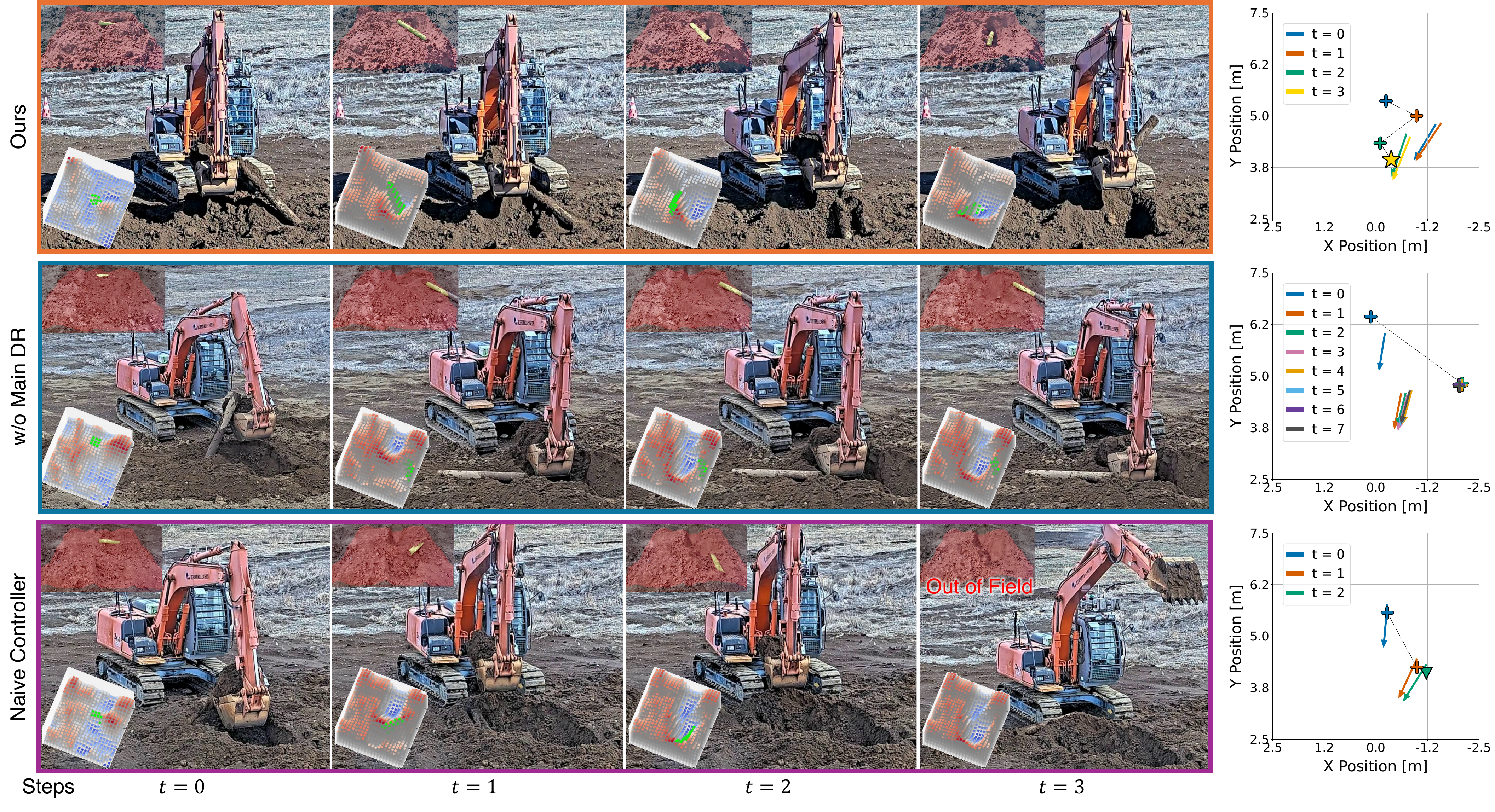}
    \caption{
        Analysis of policy behavior. Snapshots illustrate the environment during task execution. The overlaid 3D heatmap visualizes the observed depth map $I^{\mathrm{depth}}$, where the height of each pixel corresponds to its depth value. Green regions indicate obstacle locations derived from the segmentation map $I^{\mathrm{label}}$, while remaining colored regions represent depth values encoded according to $I^{\mathrm{depth}}$. Real-world images correspond to the moment immediately after executing each action at waypoint $\mathbf{p}_5$.
       Plot on the right shows policy action sequence. Arrows represent actions, where arrow origin corresponds to policy-generated XY position and arrow length indicates drag distance. Plus marker denotes obstacle centroid computed from segmentation map. A star marker indicates successful obstacle removal in subsequent action, whereas a triangle marker indicates failure. Failure is defined as the obstacle becoming unobservable by the perception camera. Obstacle used in this analysis is an I-shape object evaluated in real-world environment.
    }
    \label{fig:real_trajectory}
\end{figure*}

\subsection{Training-Time Efficiency of Burial-Conditioned Curriculum}
\label{sec:calculation_time_comparison}

    \subsubsection{Settings}
        The substantial calculation time required for sample collection in a particle simulation may hinder policy learning within practical time constraints. In this section, we measure the actual calculation time of the particle simulation and evaluate its impact on training efficiency.
        To analyze the relationship between particle quantity and calculation time, we measure the calculation time per simulation step under different particle counts corresponding to C1, C2, and C3. We then examine whether the proposed burial-conditioned curriculum improves training-time efficiency compared to fixed-particle training by measuring the wall-clock time required for the proposed framework (``Ours'') to reach its peak performance.

    \subsubsection{Results}
        \figref{fig:learning_design} (b) shows the relationship between particle quantity and calculation time per simulation step. As the number of particles increases from C1 to C3, calculation time increases significantly, indicating that higher particle counts substantially increase computational burden. In particular, the calculation time in C3 is approximately 20 times that of C1. \figref{fig:learning_design} (b) also shows the wall-clock training time required for ``Ours'' to reach peak performance. Training conducted directly in C3 requires substantially longer training time due to the high computational cost per environment interaction. In contrast, the burial-conditioned curriculum enables early learning in computationally cheaper settings (C1 and C2) before transitioning to C3. As a result, ``Ours'' requires approximately \SI{50}{\%} of the training time compared to ``Only C3.'' These results indicate that progressive particle scaling improves overall training-time efficiency by reducing computational burden during the early learning stages.

\subsection{Policy Transferability from Training Simulator to Evaluation Simulator and to Real-World Environment}
\label{sec:policy_transferability}

    \subsubsection{Settings}
       This section evaluates whether the policy trained in Isaac Gym transfers to both the real-world environment and OperaSim \cite{opera}, an evaluation simulator based on AGX Dynamics that provides more accurate soil modeling. Because real-world experiments are restricted by safety limits on obstacle size and weight, OperaSim is additionally used to assess transferability under more challenging conditions. We compare the fully randomized policy (``Ours'') with ``w/o main DR,'' which randomizes only task-irrelevant parameters, and with the centroid-based ``Naive Controller'' to examine the roles of domain randomization and policy learning in transfer performance.

    \subsubsection{Results}
        \figref{fig:task_achievement_real} shows that ``Ours'' achieves success rates above \SI{90}{\%} in both the evaluation simulator (OperaSim) and the real-world environment, whereas ``w/o main DR'' and ``Naive Controller'' remain below \SI{20}{\%}, demonstrating the proposed policy's successful sim-to-real transfer across simulators and to the real excavator. The qualitative behaviors in \figref{fig:real_trajectory} explain this gap: ``Ours'' progressively excavates around the obstacle and brings it into a removable state, while ``Naive Controller'' repeatedly scoops near the estimated centroid and often pushes the obstacle out of the task region. Although ``w/o main DR'' sometimes starts near the obstacle, the limited diversity of state--action transitions it experiences leads to repetitive excavation at similar locations and eventual failure.

\subsection{Comparison with Human Experts}
\label{sec:task_completion_time_comparison}

    \subsubsection{Settings}
       In this section, we compare the task completion time of the learned obstacle-removal policy with that of human experts in the real-world environment to assess the practical feasibility of autonomous obstacle removal and to analyze the remaining efficiency gap. We also decompose the task execution process to identify which stages dominate the total duration and to grasp how they differ from human operation. For the human baseline, a licensed excavator operator with more than one year of professional experience performs the same obstacle removal task.

    \subsubsection{Results}
\begin{figure}[t]
    \centering
    \includegraphics[width=0.99\columnwidth]{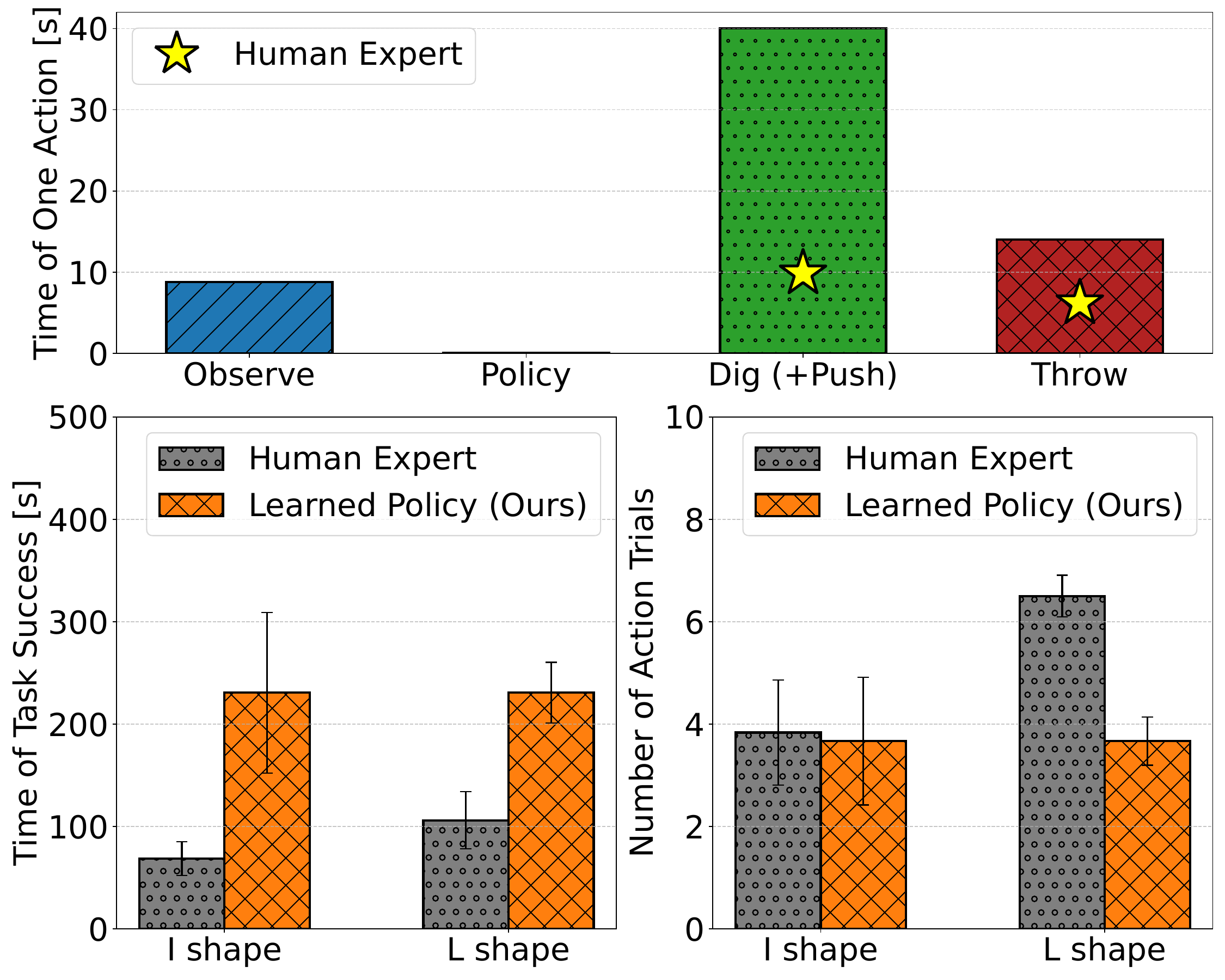}
    \caption{
        Comparison of task completion time between learned policy and human experts. Two obstacle types, I-shape and L-shape, are used under the same settings as in \figref{fig:task_achievement_real}. ``Time of One Action'' decomposes execution time of a single learned action into four components: ``Observe,'' the time from RGB-D image acquisition to conversion into observation $s$; ``Policy,'' the inference time required to generate action $a$; ``Dig (+Push),'' the time required to execute excavation; and ``Throw,'' the time required to dispose of excavated material at designated location. Star markers indicate average time of human experts, where pushing motions (frequently performed by human operators) are included in ``Dig (+Push).'' ``Time of Task Success'' represents total time for obstacle removal, and ``Number of Action Trials'' denotes number of excavation attempts, computed as $(\text{Dig}+\text{Throw})/2$. Each bar represents mean with standard deviation. For both learned policy and human operators, the task starts with bucket lifted above the ground, and total time required to excavate the obstacle and move it to predefined target location is measured.
    }
    \label{fig:compare_time}
\end{figure}

        The experimental results are shown in \figref{fig:compare_time}. Human experts complete the obstacle removal task in approximately \SI{100}{s} on average, whereas the learned policy requires approximately \SI{230}{s}. This result indicates that, although the proposed policy already permits fully autonomous obstacle removal in the real world, its execution efficiency remains below that of skilled operators. To better understand this gap, we decompose the task into per-action duration and action count. The ``Dig'' phase dominates the execution time of the learned policy, requiring approximately \SI{40}{s} per excavation, while human experts complete excavation in a substantially shorter time. Similarly, the learned policy takes approximately \SI{14}{s} in the ``Throw'' phase, about twice as long as human operators. In contrast, human experts perform a greater number of shorter excavation actions, suggesting that frequent motion switching enables them to reduce overall task completion time.

        The longer ``Dig'' duration of the learned policy mainly stems from its inability to detect obstacle drop events during excavation. To prevent the obstacle from falling out of the bucket, the generated trajectory includes many intermediate waypoints, resulting in cautious bucket motion. As shown in \tabref{table:parameter_setting}, the trajectory includes 41 waypoints, and reducing this number could shorten execution time by approximately one second per waypoint; however, this would also degrade excavation stability, which presents a trade-off between speed and robustness. Similarly, during the ``Throw'' phase, the learned policy disposes of soil outside of the task field to avoid re-deposition onto the obstacle. While this enhances task reliability, shortening the dumping distance could reduce execution time. 
        Although the primary objective of this paper is reliable obstacle removal rather than speed optimization, this comparison provides an important practical perspective: Despite the execution-time gap, the proposed framework consistently completes the task autonomously without human intervention under varying obstacle conditions, while the remaining gap from expert operators can mainly be attributed to conservative motion execution. This finding clarifies a specific direction  that can be taken for future improvement.

\subsection{Application to Practical Excavation Scenarios}
\label{sec:practical_excavation_applications}

    This section demonstrates the proposed obstacle-removal policy in two practical tasks. \textbf{Trench Excavation with Buried Obstacles:} We integrate the policy into a trench excavation task using an excavator and construct a trench of \SI{3}{m} width, \SI{10}{m} length, and \SI{1}{m} depth. Soil excavation is generated by the same four-parameter trajectory representation as in the policy execution. During each cycle, an image-based detector determines whether the bucket contains an obstacle; if so, the contents are discarded to the right, otherwise to the left. The system alternates between trench excavation and obstacle removal, then autonomously repositions forward until the full \SI{10}{m} trench length is completed. \textbf{Obstacle Removal under Rainy Condition:} We also demonstrate obstacle removal under rainy outdoor conditions using the same I-shaped obstacle as in \figref{fig:task_achievement_real}. The learned policy (``Ours'') is executed without additional tuning. Representative scenes are shown in \figref{fig:application}, and detailed motions are provided in the supplementary videos.

\begin{figure}[t]
    \centering
    \includegraphics[width=0.99\columnwidth]{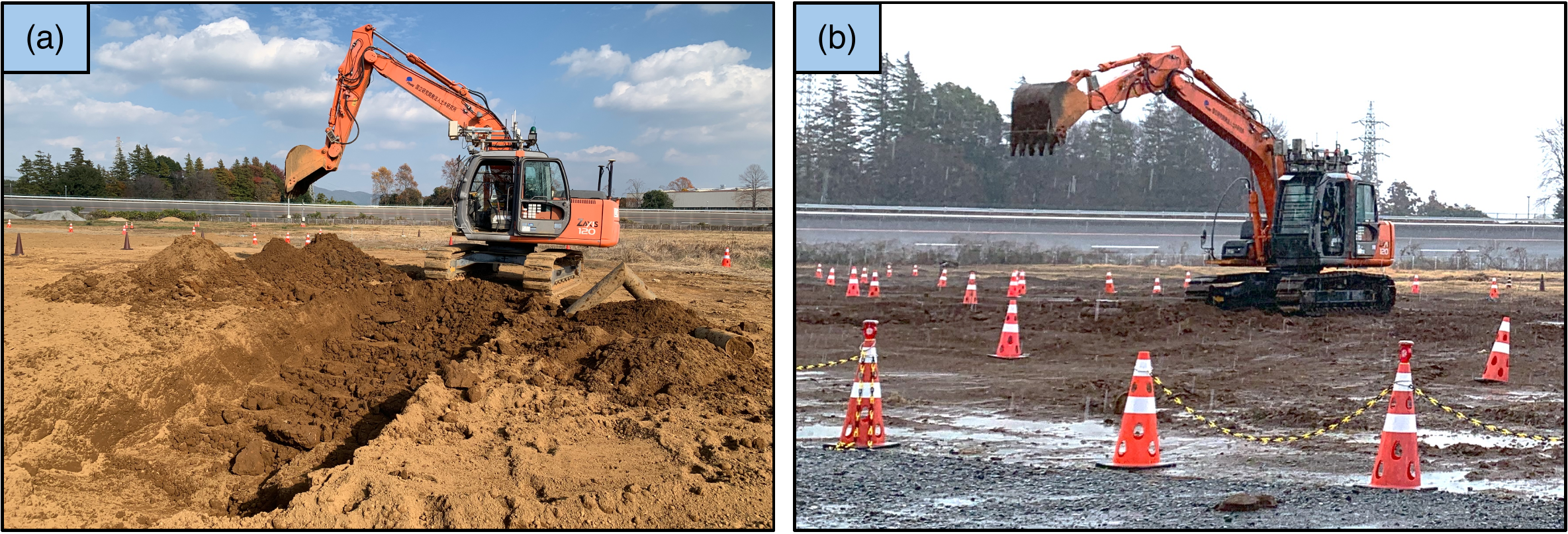}
    \caption{
        Application demonstrations of proposed obstacle-removal policy.
        \textbf{(a)} Trench excavation with buried obstacles after integrating the policy into a trench excavation plan.
        \textbf{(b)} Obstacle removal under rainy outdoor conditions.
    }
    \label{fig:application}
\end{figure}

\section{Discussions}
\label{s:dis}

    This section discusses the current scope and remaining limitations of the proposed framework and highlights several extensions that are important for broader practical deployment. In particular, we focus on issues related to obstacle properties, curriculum design, terrain conditions, integrated mobility, perception, and observation design.

    \subsection{Extending Framework to Obstacles That Deform or Break During Excavation}

    Handling obstacles that may deform or break during excavation requires extending the current rigid-body framework, since deformable-object manipulation and realistic simulation of deformation and fracture remain active research topics in robotics and simulation \cite{sanchez_deformable_survey,diffsrl,rl_tool_life}. In practice, obstacles such as wood may crack under bucket forces, and excavation may even exploit this property by first breaking the obstacle into smaller pieces and then more easily removing them. Addressing such cases would require replacing the current rigid obstacle model with one that can represent deformation and fracture, potentially creating policies that make use of material breakage during removal. Developing such models, however, is beyond the scope of this study and remains future work.

    \subsection{Why Starting From Easier Stages Remains More Time-Efficient Than Direct Training in C3}

    While this paper demonstrates the effectiveness of beginning training from C1, it is natural to ask whether the policy could instead be learned directly in C3. The key advantage of the burial-conditioned curriculum is that early learning is carried out in computationally inexpensive environments. As a result, the wall-clock benefit of the curriculum remains even if sample efficiency improves, since that improvement would benefit both the proposed method and direct training in C3 \cite{CRL,curriculum_learning_Visuomotor}.

    \subsection{Additional Stability Challenges in Extending Policy to Sloped Terrain}

    The obstacle-removal policy developed in this paper assumes operation on flat terrain and thus does not address sloped environments. However, excavation is often performed on slopes or uneven ground in practical construction and disaster-response scenarios, making this extension an important direction for future research \cite{heap_walking_excavator,ito_machine_instability}. Applying the proposed approach to such settings would require training in simulation environments where both the terrain and the excavator are inclined. Under these conditions, bucket reaction forces may cause the machine body to lift or slide down the slope, introducing additional stability constraints that are absent on flat ground. Addressing these effects would require further extension of the framework, so this issue is left for future work.

    \subsection{Integrating Mobility for practical deployment}

    In this study, mobility is not considered in the control policy, except in the trench excavation demonstration where it is used only to move to another working area at the next trench location. In practical construction and disaster-response scenarios, however, autonomous operation requires integrated control of excavation and mobility \cite{garrett_itamp_review,robo_embankment}. For example, when an obstacle is too close to the machine body for excavation, the excavator would need to be repositioned to allow the task to continue. Extending the framework to such mobility-aware planning is thus an important direction for autonomous construction robotics. This extension also raises safety issues, since communication loss in the mobility system may cause unintended movement or collisions. Developing safe experimental environments and validation procedures for this setting remains future work.

    \subsection{Difficulty of Scaling Supervised Perception Pipeline to Diverse Obstacles}

    In this paper, obstacle detection in the real world uses RGB images and a supervised perception pipeline composed of Faster R-CNN and GrabCut refinement to obtain accurate object boundaries. The detector is trained on manually annotated real-world images collected under various obstacle configurations. Extending this approach to more diverse obstacles, such as rocks, fallen trees, and unknown objects, would require substantial labeled data and thus high annotation cost, which is consistent with the broader limitations of supervised and open-world object detection \cite{faster_rcnn,owod_cvpr2021}. Because perception is not the primary focus of this work, we adopt this pipeline here for reliable detection. Accordingly, developing more generalizable perception methods for diverse real-world obstacles remains future work.

    \subsection{Trade-offs Among Continuous Visual Feedback, Occlusion Robustness, and Sample Efficiency}

    In this paper, the policy observes the environment only when the bucket is lifted, and then it generates a single excavation trajectory from that visual input. Continuous observation during excavation could improve performance, but we do not adopt it because the bucket occludes the ground and obstacles, which increases observation noise and degrades state estimation; furthermore, severe occlusion remains challenging even in conventional visual recognition pipelines \cite{occluded_rcnn,deep_visual_foresight}. Reproducing such occlusions accurately in simulation is also difficult and may enlarge the gap from reality. In addition, more frequent observations increase input dimensionality and may reduce sample efficiency, offsetting the advantage of trajectory parameterization. Continuous observation is thus a promising but challenging extension that must be left for future work.

\section{Conclusion}
\label{s:con}

    This paper presented a sim-to-real framework for autonomous obstacle removal with an excavator. The proposed framework combines a trajectory-level action representation, integrated depth and obstacle-label observations, and a burial-conditioned curriculum to address both the learnability and computational cost of obstacle removal in particle simulation. Results of comparative experiments show that the proposed method learns a high-performing policy more reliably than alternative training strategies, and findings from ablation studies confirm that both the action representation and the observation design contribute substantially to task performance. The curriculum reduced wall-clock training time while maintaining good overall performance, and transfer experiments demonstrated robust obstacle removal in both an additional simulator used for evaluation and a real excavator. A comparison with human experts further clarified that the remaining limitation is predominantly execution efficiency rather than task feasibility, thus indicating a concrete direction for future improvement.

\appendices

\section{Detailed Definition of Trajectory Parameterization}
    \label{sec:trajectory_parameterization_details}

    This appendix supplements the action design in \chapref{action_design}. The policy outputs four parameters, $a=[X,Y,D,L]^\top$, and a deterministic trajectory generator converts these into bucket motion that can be executed in both simulation and the real system.

    Once $(X, Y)$ is given, a vertical excavation plane is defined by the excavator base and the selected excavation position. The penetration point on this plane is defined as
    \begin{equation}
        \mathbf{p}_2 = [X, Y, -D]^\top.
    \end{equation}
    Here, the excavation direction is determined by the swing angle $\theta$, defined as $\theta = \arctan\left(\frac{Y}{X}\right)$, and the quadrant is chosen consistently with $(X, Y)$. This yields the horizontal unit vector
    \begin{equation}
        \mathbf{d} = [\cos\theta, \sin\theta, 0]^\top.
    \end{equation}
    Using this direction, the digging endpoint is defined as
    \begin{equation}
        \mathbf{p}_3 = \mathbf{p}_2 + L \mathbf{d}.
    \end{equation}
    Additional control points are then introduced for the approach, lift-out, and transfer phases:
    \begin{equation}
        \mathbf{p}_1 = \mathbf{p}_2 + \alpha \mathbf{d} + [0, 0, h]^\top,
    \end{equation}
    \begin{equation}
        \mathbf{p}_4 = \mathbf{p}_3 - \beta \mathbf{d} + [0, 0, h']^\top,
    \end{equation}
    \begin{equation}
        \mathbf{p}_5 = \mathbf{p}_4 - \gamma \mathbf{d} + [0, 0, H]^\top,
    \end{equation}
    where $\alpha$, $\beta$, and $\gamma$ are predefined offsets along the excavation direction, and $h$, $h'$, and $H$ are predefined vertical offsets. Here, $\mathbf{p}_1$ is the approach point, $\mathbf{p}_2$ is the penetration point, $\mathbf{p}_3$ is the end of dragging, $\mathbf{p}_4$ is the lift-out point, and $\mathbf{p}_5$ is the removal point.

\begin{figure}[t]
    \centering
    \includegraphics[width=0.95\columnwidth]{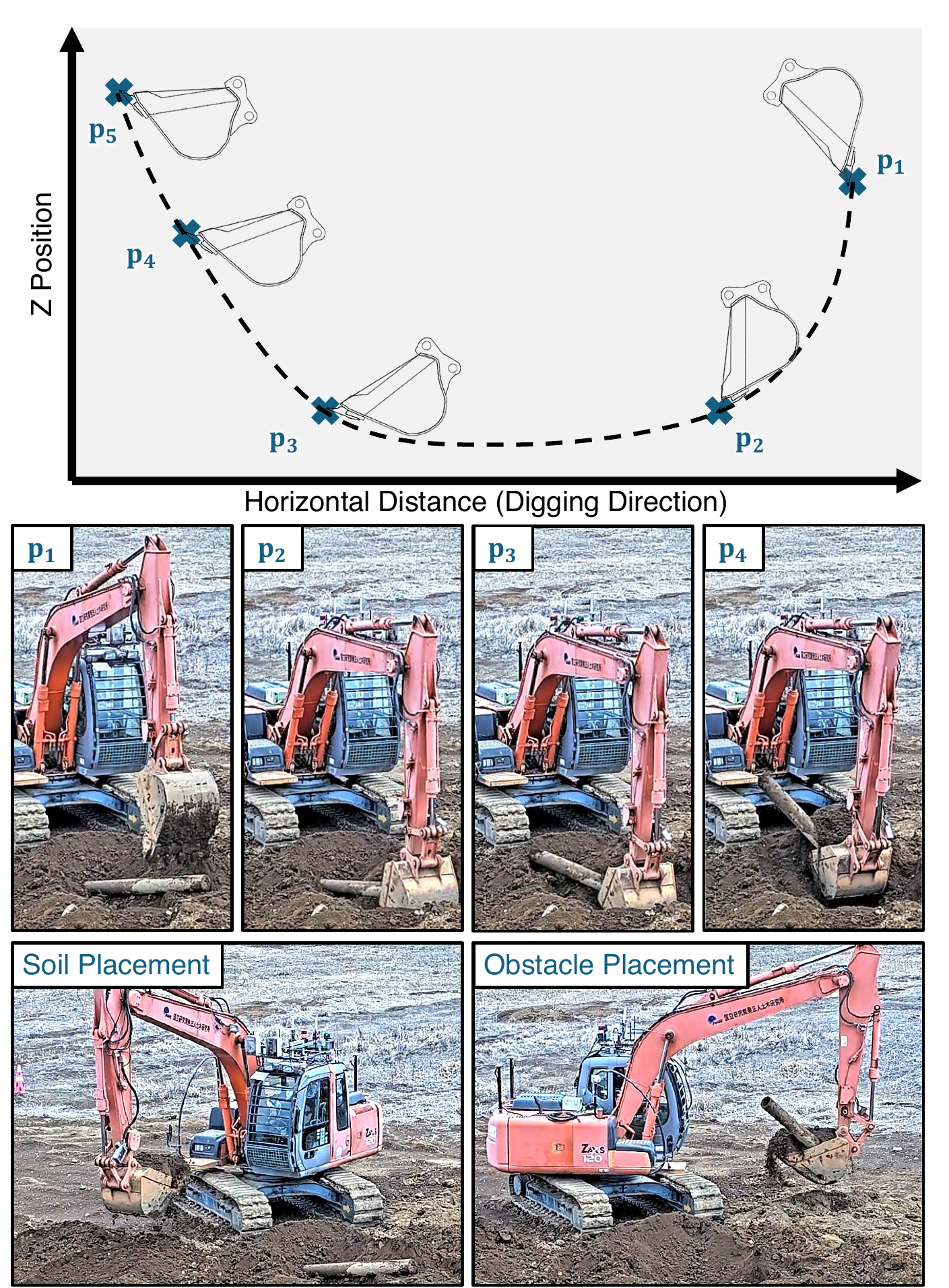}
    \caption{
        Detailed definition of trajectory parameterization. Action parameters determine five control points $\mathbf{p}_1$ to $\mathbf{p}_5$ for the approach, penetration, dragging, lift-out, and removal phases. After reaching $\mathbf{p}_5$, the bucket contents are discharged either to the soil-placement point for soil or to the obstacle-placement point for obstacles.
    }
    \label{fig:path_design}
\end{figure}

    As illustrated in \figref{fig:path_design}, the excavation phase is generated by cubic spline interpolation \cite{splines} over $\mathbf{p}_1$, $\mathbf{p}_2$, and $\mathbf{p}_3$:
    \begin{equation}
        \mathbf{x}(\tau) = \mathrm{Spline}(\mathbf{p}_1, \mathbf{p}_2, \mathbf{p}_3), \quad \tau \in [0,1].
    \end{equation}
    The bucket orientation is aligned with the trajectory tangent,
    \begin{equation}
        \mathbf{t}(\tau) = \frac{d\mathbf{x}(\tau)}{d\tau}, \quad \mathbf{R}(\tau) = \mathrm{Align}(\mathbf{t}(\tau)),
    \end{equation}
    and the resulting motion is discretized into waypoints and executed sequentially by the low-level controller. After excavation, the bucket is lifted to $\mathbf{p}_5$, where its contents are checked, and a dumping motion to either the soil-placement point or the obstacle-placement point is selected according to whether the bucket contains soil only or an obstacle.

\section{Detailed Real-World Observation and Action Execution}
    \label{sec:real_execution_details}

    This appendix supplements \chapref{sec:proposed_real_execution}. The real system preserves the observation--action interface used in simulation so that the learned policy can be deployed without additional fine-tuning. The excavator is controlled through the same primary joints as in simulation---swing, boom, arm, and bucket---with the joint states measured by onboard absolute encoders and bucket orientation estimated by an inertial measurement unit (IMU). An RGB-D camera (Intel RealSense) observes the task field, and observations are acquired only when the bucket is lifted to match the timing used in simulation.

    \subsubsection{Observation Construction}
        The real system provides the same two observation channels as in simulation, namely the terrain representation $I^{\mathrm{depth}}$ and the obstacle-region representation $I^{\mathrm{label}}$. The depth observation $I^{\mathrm{depth}}$ is generated by transforming the RGB-D point cloud from the camera frame to the global robot frame, filtering it within a predefined 3D workspace, and projecting the remaining points onto a 2D grid on the $(X, Y)$ plane, where each cell stores the maximum height value. Missing cells due to occlusion are filled using a scattered-data interpolation method \cite{interpolation}.

        Obstacle information is obtained by a supervised perception pipeline. The RGB image is processed by multi-scale Retinex \cite{multiscale_retinex} and a bilateral filter \cite{bilateral_filter}, object proposals are detected by Faster R-CNN \cite{faster_rcnn}, and the detected regions are refined by GrabCut \cite{gradcut} to obtain obstacle masks. The resulting segmentation filters the point cloud, and the extracted obstacle points are projected onto the same grid to generate the binary label image $I^{\mathrm{label}}$. Finally, both $I^{\mathrm{depth}}$ and $I^{\mathrm{label}}$ are represented on a fixed-resolution grid and downsampled to match the policy input used in simulation.

    \subsubsection{Low-Level Action Execution}
        The policy output is the parameterized excavation command $a = [X, Y, D, L]^\top$, which the real system converts into the same trajectory-level motion structure used in simulation. A deterministic trajectory generator produces intermediate waypoints from the commanded parameters, and each waypoint is converted into joint targets through inverse kinematics. These targets are executed by onboard joint controllers with a PID-based scheme, where each waypoint is tracked for a predefined actuation duration. To handle execution variability, a timeout mechanism advances to the next waypoint if the target configuration is not reached within the specified duration. This preserves the action interface learned in simulation while allowing robust execution on the real excavator.

\begin{table}[t]
    \vspace{1.mm}
    \begin{center}
    \caption
    {
        Range of randomized parameters in the simulator.
        \label{table:domain-parameter}
    }
    \footnotesize
        \begin{tabular}{lccc}
            \toprule
                Parameter & w/o DR & min & max \\
            \midrule
                Soil particle friction coefficient $[\cdot]$ & 0.7 & 0.4 & 1 \\
                Soil particle mass $[\textrm{kg}]$ & 20 & 10 & 30 \\
                Soil particle edge length $[\textrm{m}^3]$ & 0.2 & 0.16 & 0.24 \\
                Soil base height bias $[\textrm{m}]$ & 0 & -0.5 & 0.5 \\
                Cover soil cluster number $[\cdot]$ & 0 & 5 & 15 \\
                Cover soil cluster particle amount $[\cdot]$ & 0 & $3^3$ & $5^3$ \\
                Cover soil cluster position bias (XY) $[\textrm{m}]$ & 0 & -2 & 2 \\
                Camera position bias (XYZ) $[\textrm{m}]$ & 0 & -0.1 & 0.1 \\
                Camera angle bias (XYZ) $[\textrm{degree}]$ & 0 & -10 & 10 \\
                Camera depth noise $[\textrm{m}]$ & 0 & -0.1 & 0.1 \\
                Camera depth bias $[\textrm{m}]$ & 0 & -0.3 & 0.3 \\
                Obstacle initial position bias (XY) $[\textrm{m}]$ & 0 & -1.5 & 1.5 \\
                Obstacle friction coefficient $[\cdot]$ & 0.5 & 0.1 & 1 \\
                Obstacle part box mass $[\textrm{kg}]$ & 10 & 5 & 20 \\
                Obstacle part box number $[\cdot]$ & 1 & 1 & 8 \\
                Obstacle part box edge length $[\textrm{m}^3]$ & 0.3 & 0.24 & 0.36 \\
            \bottomrule
        \end{tabular}
    \end{center}
\end{table}

\section{Detailed Settings for Simulation and Domain Randomization}
    \label{sec:simulation_dr_details}

    This appendix supplements \chapref{sec:proposed_simulation_learning}. The simulator provides the same observation--action interface as that used by the real system. A virtual overhead camera is implemented using the depth-rendering and semantic-segmentation functions of Isaac Gym so that the observation matches the real perception pipeline, producing a depth image $I^{\mathrm{depth}}$ and an obstacle label image $I^{\mathrm{label}}$. The simulator also reproduces the PID-based control structure used on the real excavator so that each trajectory-level action is executed through the same low-level interface in both environments.

    Domain randomization is applied to three groups of factors. First, soil and terrain properties are randomized by varying the size, mass, and friction of coarse cubic particles as well as the number, position, and height of particle clusters within the task field. Second, perception robustness is improved by perturbing camera pose and applying noise and bias to depth observations, while both simulation and real observations are downsampled to reduce sensitivity to fine-grained discrepancies. Third, obstacles are randomized by their properties of friction, mass, initial pose, and geometry.

    To generate diverse obstacle shapes, the simulator incrementally connects cuboid primitives. A base cuboid with size $\mathbf{s}_{\mathrm{base}}$ is placed first, and a set of candidate connection positions $\{\mathbf{p}_i\}$ is maintained. At each step, one candidate is selected and a new cuboid is attached along an axis $a \in \{x,y,z\}$ with a direction bias chosen to suppress excessive elongation. The size of each cuboid is perturbed as $\mathbf{s}_i = \mathbf{s}_{\mathrm{base}} \odot (1 + \boldsymbol{\epsilon})$, where $\boldsymbol{\epsilon} \sim \mathcal{U}(-\alpha, \alpha)$. If a generated position exceeds a distance threshold from the centroid, it is corrected inwardly so that the final merged obstacle mesh remains spatially compact. The randomized parameter ranges used in the simulator are summarized in \tabref{table:domain-parameter}.

\begin{table}[t]
    \caption{
            Shared experimental settings and parameters.
            \label{table:parameter_setting}
    }
    \begin{center}
        \footnotesize
        \begin{tabular}{@{}p{7cm}p{1.5cm}@{}}
            \toprule
            \textbf{Parameter} & \textbf{Value}  \\ 
            \midrule
            PID timeout [s] & 1 \\
            PID frequency [Hz] & 100 \\
            Waypoint number ($\mathbf{p}_1$ to $\mathbf{p}_2$) & 13 \\
            Waypoint number ($\mathbf{p}_2$ to $\mathbf{p}_3$) & 12 \\
            Waypoint number ($\mathbf{p}_3$ to $\mathbf{p}_4$) & 8 \\
            Waypoint number ($\mathbf{p}_4$ to $\mathbf{p}_5$) & 8 \\
            Start/end height above ground surface of Trj. baseline [m] & 1 \\
            Total waypoint number for Trj. baseline & 30 \\
            Approach offset $\alpha$ along excavation direction for $\mathbf{p}_1$ [m] & 0.6 \\ 
            Lift-out offset $\beta$ opposite to excavation direction for $\mathbf{p}_4$ [m] & 0.5 \\ 
            Removal offset $\gamma$ opposite to excavation direction for $\mathbf{p}_5$ [m] & 0.2 \\
            Dig-in height $h$ [m] & $D + 0.5$ \\ 
            Dig-out height $h'$ [m] & $D$ \\
            Dig-out height $H$ [m] & 1.2 \\
            Bucket angle at $\mathbf{p}_4$ [deg] & 160 \\
            Bucket angle at $\mathbf{p}_5$ [deg] & 170 \\
            Action range (X) & 4.3 -- 6.1 \\
            Action range (Y) & -1.5 -- 1.5 \\
            Action range (D) & -1.5 -- -0.5 \\
            Action range (L) & 0.5 -- 1.5 \\
            Camera observation resolution & $42 \times 42$ \\
            Camera field range in machine-centered $x$ direction [m] & 2.5 -- 7.5 \\
            Camera field range in machine-centered $y$ direction [m] & -2.5 -- 2.5 \\
            \midrule
            Number of parallel environments (C1) & 64 \\
            Number of parallel environments (C2) & 32 \\
            Number of parallel environments (C3) & 16 \\
            Target success rate for curriculum transition & 0.95 \\
            Number of steps per episode & 8 \\
            \midrule
            RL algorithm & REDQ \cite{REDQ} \\
            Number of critics & 10 \\
            Number of randomly sampled critics for target estimation & 2 \\
            Policy updates per step & 1 \\
            Critic updates per step & 5 \\
            Optimizer & Adam \\
            Learning rate & $3 \times 10^{-4}$ \\
            Batch size & 256 \\
            Replay buffer size & $10^6$ \\
            Success reward scale $\omega_{\mathrm{success}}$ & 1 \\
            Distance reward scale $\omega_{\mathrm{distance}}$ & 0.2 \\
            \bottomrule
        \end{tabular}
    \end{center}
\end{table}

\begin{figure*}[t]
    \centering
    \includegraphics[width=1.95\columnwidth]{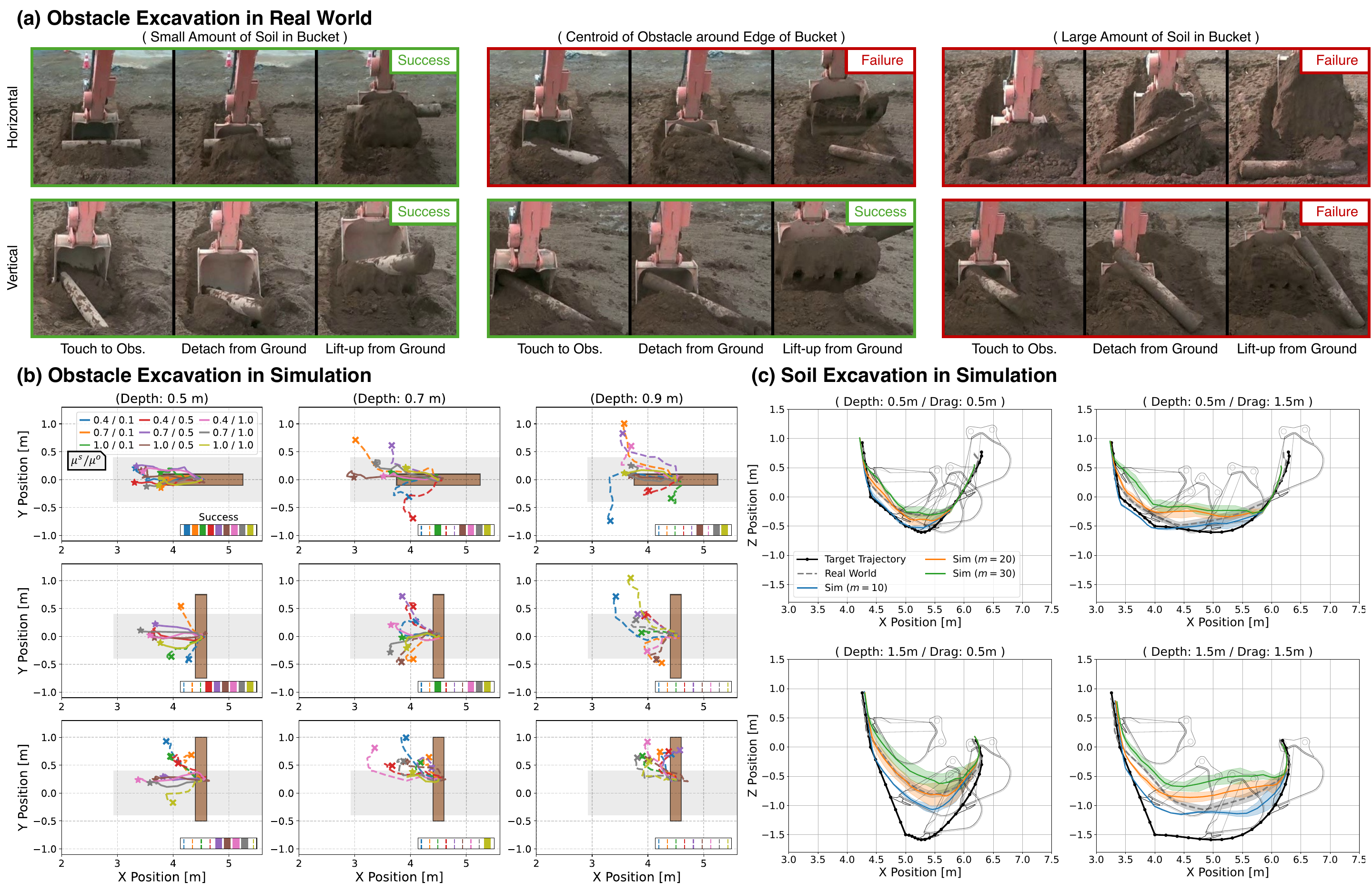}
    \caption{
        Qualitative comparison of predefined excavation behaviors between real-world environment and simulation environment.
        \textbf{(a)} Real-world obstacle excavation for different excavation positions and soil amounts; each case shows initial contact, bucket lift-off, and scene after lifting.
        \textbf{(b)} Simulated obstacle excavation for different friction settings, where $\mu^s$ and $\mu^o$ denote soil and obstacle friction coefficients, respectively; gray region indicates bucket sweep, and star/cross markers denote success/failure. Other inputs are fixed to $X = 5$, $Y = 0$, $L = 1.5$, and soil mass $m = 10$.
        \textbf{(c)} Bucket trajectories during soil excavation. Real-world trajectories are compared with simulated trajectories under different soil masses $m$ and $\mu^s \in [0.4, 1.0]$; curves show mean and standard deviation, with $X = 5.2$ and $Y = 0$ fixed.
    }
    \label{fig:ex_DR_param_whole}
\end{figure*}

\section{Shared Experimental Settings and Parameters}
\label{sec:experimental_parameters}

    This appendix serves as a reference for the \textit{Common Settings} subsection in the main text. It consolidates the numerical settings shared by simulation, real-world execution, and policy learning so that the main experimental sections can focus on the differences among evaluations. \tabref{table:parameter_setting} is organized into three groups. The first group lists the parameters of the common trajectory generator and observation setup, including the waypoint discretization between $\mathbf{p}_1$ and $\mathbf{p}_5$, the fixed geometric offsets, the action ranges, and the camera observation region. The policy outputs only the four action parameters $[X,Y,D,L]^\top$; the remaining trajectory quantities are fixed shared settings, and their geometric roles are detailed in \apperef{sec:trajectory_parameterization_details}. The second group summarizes the burial-conditioned curriculum settings shared across training runs, i.e., the number of parallel environments in C1--C3, the success-rate threshold for curriculum transition, and the episode horizon. The final group lists the REDQ hyperparameters used in all learning-based methods.
    Policy learning is run on a workstation with an Intel Core i9-9900X CPU and an NVIDIA GeForce RTX 4090 GPU.

\section{Detailed Definition of Waypoint-Based Trajectory Baseline}
    \label{sec:trajectory_baseline_details}

    This appendix supplements the description of ``Only C3 (Trj.)'' in \chapref{sec:comparing_methods}. As a waypoint-based baseline, we consider direct trajectory learning (``Trj.''), inspired by waypoint-based excavation trajectory representations \cite{soil_formula_2} and adapted here to our obstacle-removal setting on the ground. The baseline replaces the proposed four-parameter action with a direct waypoint output while keeping the observation space, reward design, learning algorithm, and low-level controller unchanged.

    In this baseline, the policy first outputs the horizontal coordinates $(X, Y)$ of the start point, while the start height $Z$ is fixed to a predefined height above the ground surface. It then outputs $n$ intermediate points, where $n$ denotes the number of intermediate points in ``Trj. ($n$).'' Each intermediate point is represented relative to the previous point by three values that specify the excavation depth, drag distance, and bucket rotation, resulting in an action dimension of $2 + 3n$.

    To ensure that the bucket exits the soil after excavation, an additional terminal point is appended. This terminal point shares the horizontal coordinates of the last intermediate point, while its height is reset to the same predefined height as the start point. The start point, the learned intermediate points, and the terminal point are then converted into a continuous trajectory using the same interpolation function as that in the proposed method, with a fixed total of 30 interpolated waypoints distributed across segments. The interpolated bucket position and angle are executed by the same PID-based low-level controller as that used in the proposed parameterized-action method.

\section{Qualitative Comparison of Excavation Behaviors Between Simulation and Real World}
\label{sec:simulation_fidelity_analysis}

    \subsubsection{Settings}
        To contextualize the training environment, we qualitatively compare representative excavation behaviors in simulation with those of the real excavator under the same control commands. The goal is to assess whether key excavation dynamics and the range induced by domain randomization are qualitatively consistent with real execution, rather than to obtain an exact numerical agreement.

    \subsubsection{Results}
        \figref{fig:ex_DR_param_whole}(a) shows that removal is more likely when the bucket interior is relatively empty and the obstacle centroid falls inside the bucket, whereas failures occur more often when the bucket is already full of soil or the centroid is outside the bucket. For the I-shaped obstacle, a vertical orientation is more favorable because the centroid more easily enters the bucket region.

        \figref{fig:ex_DR_param_whole}(b) shows the same tendency in simulation: Deeper excavation increases soil accumulation and can cause lateral slip of the obstacle, whereas vertically oriented obstacles are easier to capture. \figref{fig:ex_DR_param_whole}(c) shows that varying the soil particle mass produces simulated trajectories that span those observed in the real world; in both environments, larger soil resistance prevents the bucket from reaching the commanded depth. Overall, the simulator reproduces the interaction patterns most relevant to interpreting the policy-learning results, although the comparison remains qualitative rather than a strict validation of physical accuracy.

\section{Analysis of Trajectory Parameterization for Sample Efficiency}
    \label{sec:trajectory_design_analysis}

    \subsubsection{Settings}
        We hypothesize that trajectory parameterization is sufficient for policy learning and can improve sample efficiency in comparison to directly learning detailed intermediate waypoints. In this section, we analyze the effect of trajectory parameterization on sample efficiency by comparing the proposed parameterized representation (``Ours'') with direct waypoint-based trajectory learning (``Only C3 (Trj.)''). For the waypoint-based approach, we vary the number of trajectory waypoints to evaluate the influence of trajectory resolution on learning performance and sample efficiency.

\begin{figure}[t]
    \centering
    \includegraphics[width=0.99\columnwidth]{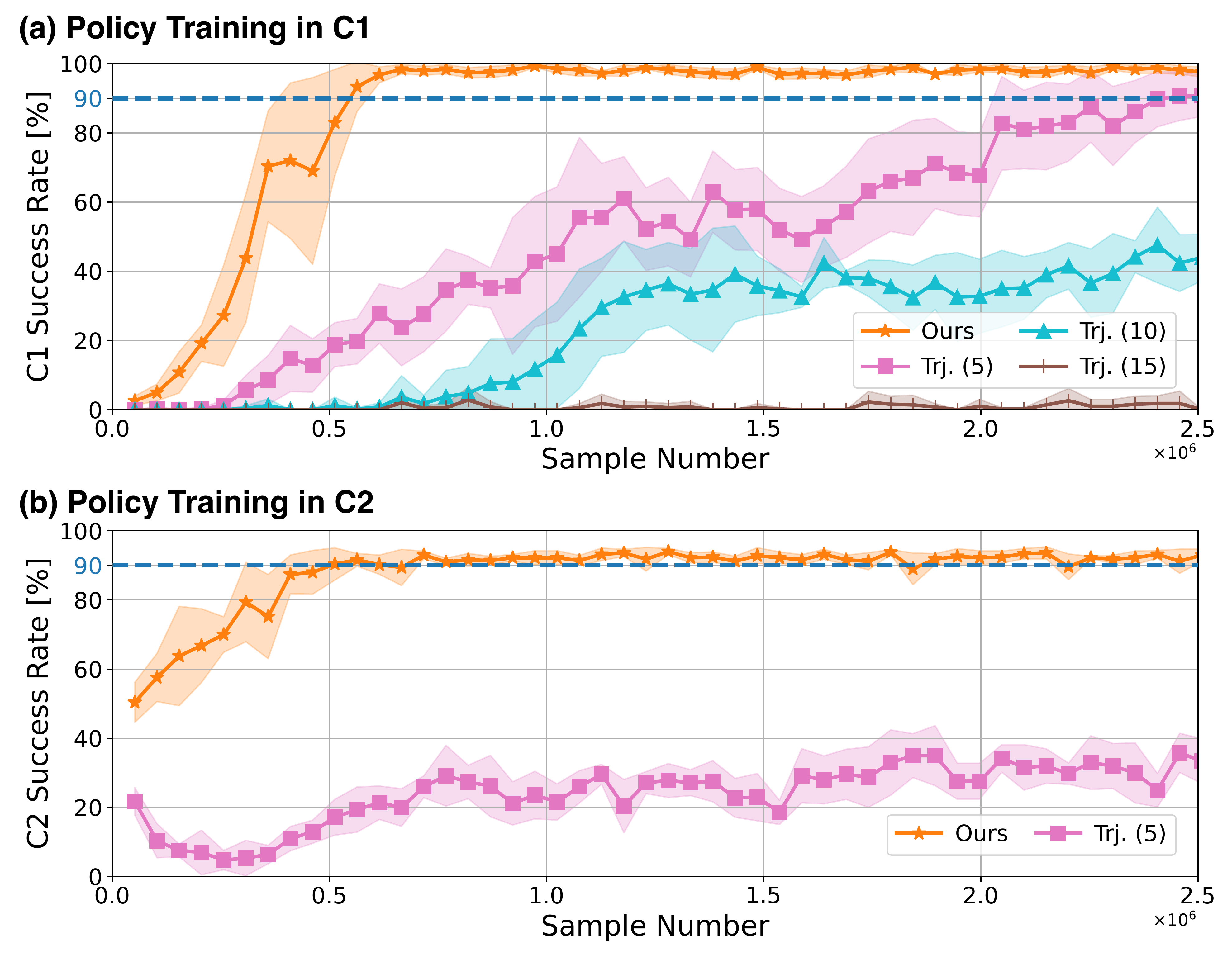}
    \caption{
        Comparison of sample efficiency between designed actions and fine time-scale actions.
        “Trj. (5)”, “Trj. (10)”, and “Trj. (15)” denote trajectories with 5, 10, and 15 intermediate waypoints.
        \textbf{(a)} Learning curves in C1, where policy is trained from scratch.
        \textbf{(b)} Learning curves in C2, where training is initialized from policy learned in C1 at the point where curriculum transition is achieved at a \SI{90}{\%} success rate.
        Note that “Trj. (10)” and “Trj. (15)” are not evaluated in C2, since policy training in C1 did not terminate within specified number of samples for these settings.
        Each curve shows mean and standard deviation over five experiments.
    }
    \label{fig:trajectory_design}
\end{figure}

\begin{figure*}[t]
    \centering
    \includegraphics[width=1.99\columnwidth]{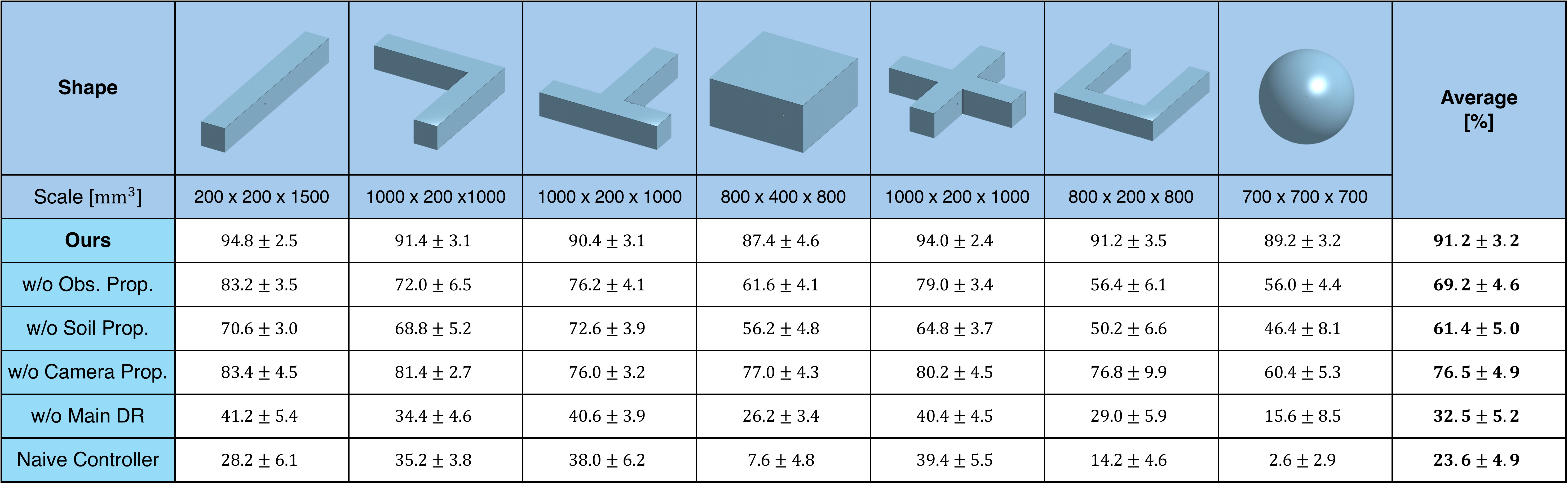}
    \caption{
        Comparison of the effects of domain randomization parameters on success rates.
        Excavation success rates for seven obstacle shapes in Curriculum 3 domain-randomized environment, evaluated over 100 trials per policy and reported as mean and standard deviation over five experiments. For centroid-based naive controller, fixed action parameters are optimized by Bayesian optimization.
        “w/o shape” denotes training with a square obstacle only, and
        ``w/o main DR'' denotes randomization only of task-irrelevant factors such as initial obstacle position.
    }
    \label{fig:task_achievement_sim}
\end{figure*}

    \subsubsection{Results}

        \figref{fig:trajectory_design} shows the learning curves for the parameterized representation (``Ours'') and the direct waypoint-based approach (``Only C3 (Trj.)'') with different waypoint counts. The proposed parameterization achieves faster convergence and higher success rates across all tested resolutions.
        Although increasing the number of waypoints improves trajectory expressiveness, it also increases action dimensionality, resulting in slower learning and reduced sample efficiency. Even with 10 or 15 waypoints, the direct waypoint-based approach fails to reach the curriculum transition threshold of \SI{90}{\%} success within the specified number of interactions. With 5 waypoints, the curriculum transition is achieved in C1 but not in C2. Interestingly, using 5 waypoints results in a parameter count comparable to that of the proposed trajectory parameterization (4 parameters), suggesting that the parameterized representation is more effective than directly learning intermediate waypoints with a similar action dimensionality. These results suggest that trajectory parameterization provides a compact yet sufficiently expressive action representation, improving exploration efficiency and overall sample efficiency.

\section{Evaluating Necessity of Policy Learning for Obstacle Removal}
\label{sec:policy_learning_necessity}

    \subsubsection{Settings}
       In this section, we evaluate whether policy learning is necessary for obstacle removal by comparing the proposed framework (``Ours'') with a centroid-based naive controller. In the proposed framework, the policy predicts the four action parameters $[X,Y,D,L]^\top$. In the naive controller, the excavation position is defined as the obstacle centroid plus a fixed insertion offset $\Delta X$ in the $X$ direction; $\Delta Y$ is set to zero, and the excavation depth $D$ and drag distance $L$ are also fixed. For fairness, $D$, $L$, and $\Delta X$ are optimized using Bayesian optimization in a domain-randomized simulation environment in which 100 trials are run, using task success rate as the objective function. The best parameters are then fixed and evaluated with the same number of excavation trials as the learning-based policy.

    \subsubsection{Results}
        \figref{fig:compare_task_design} (c) shows the Bayesian optimization results used to determine the fixed action parameters of the centroid-based naive controller. The results indicate that an insertion offset $\Delta X$ of approximately \SI{1}{m} yields the highest performance. As $\Delta X$ increases, the distance between the bucket and the obstacle also increases, requiring a larger drag distance ($L$) to reach the obstacle effectively. Combinations with larger $L$ values tend to achieve higher success rates under larger $\Delta X$ settings. The best-performing parameter combination achieves a task success rate of about \SI{50}{\%}, substantially lower than that of the learning-based excavation policy, which achieves over \SI{90}{\%}. 
        Additionally, \figref{fig:task_achievement_sim} evaluates the centroid-based naive controller across various obstacle shapes. The success rate remains roughly below \SI{30}{\%}, whereas ``Ours'' consistently achieves about \SI{90}{\%}, highlighting the limited generalization capability of the naive approach.
        These results indicate that the proposed excavation policy framework achieves superior performance to that of a centroid-based naive controller, even when its fixed parameters are optimized.

\section{Evaluating Contribution of Domain Randomization Parameters}
\label{sec:domain_randomization_contribution}

    \subsubsection{Settings}
        In this section, we evaluate the contribution of each domain-randomization (DR) group through sim-to-sim ablations with various obstacle shapes. The randomized factors are divided into soil properties, obstacle properties, and camera-related parameters. In each ablation, one group is removed while the others remain randomized. As a reference, we also evaluate ``w/o main DR,'' which randomizes only task-irrelevant factors such as initial obstacle placement.
        
    \subsubsection{Results}
        \figref{fig:task_achievement_sim} shows that the full model (``Ours'') performs best across all obstacle shapes. Removing soil or obstacle-property randomization reduces success rates to below \SI{70}{\%}, indicating the importance of interaction-related physical variation. Removing camera-related randomization causes a smaller drop to approximately \SI{80}{\%}, showing a secondary but non-negligible contribution of perceptual robustness. Randomizing only task-irrelevant factors leads to substantially lower performance. Overall, these results indicate that physical DR is essential, while camera DR provides complementary gains.

\section*{Acknowledgment}
    The authors would like to thank Yosuke Matsusaka of the Public Works Research Institute for his helpful advice on simulator-related issues during the development of OperaSim.

\addtolength{\textheight}{-0cm}   


\bibliographystyle{IEEEtran}
\bibliography{bib/IEEEabrv,bib/reference}

\end{document}